%% file: paper.tex
\documentclass[10pt,journal,compsoc]{IEEEtran}

\input{vcl-shortcuts.tex}
\graphicspath{{./figures/}}

\ifCLASSOPTIONcompsoc
  \usepackage[nocompress]{cite}
\else
  \usepackage{cite}
\fi

\usepackage[pagebackref=true,breaklinks=true,letterpaper=true,bookmarks=false]{hyperref}

\begin{document}

\title{ASH: \textit{A} Modern Framework \\ for Parallel \textit{S}patial \textit{H}ashing in 3D Perception}

\author{
  Wei~Dong,
  Yixing~Lao,
  Michael~Kaess,
  and Vladlen~Koltun
}

\IEEEtitleabstractindextext{
\begin{abstract}
We present ASH, a modern and high-performance framework for parallel spatial hashing on GPU. Compared to existing GPU hash map implementations, ASH achieves higher performance, supports richer functionality, and requires fewer lines of code (LoC) when used for implementing spatially varying operations from volumetric geometry reconstruction to differentiable appearance reconstruction.
Unlike existing GPU hash maps, the ASH framework provides a versatile tensor interface, hiding low-level details from the users. In addition, by decoupling the internal hashing data structures and key-value data in buffers, we offer direct access to spatially varying data via indices, enabling seamless integration to modern libraries such as PyTorch.
To achieve this, we 1) detach stored key-value data from the low-level hash map implementation; 2) bridge the pointer-first low level data structures to index-first high-level tensor interfaces via an index heap; 3) adapt both generic and non-generic integer-only hash map implementations as backends to operate on multi-dimensional keys.
We first profile our hash map against state-of-the-art hash maps on synthetic data to show the performance gain from this architecture. We then show that ASH can consistently achieve higher performance on various large-scale 3D perception tasks with fewer LoC by showcasing several applications, including 1) point cloud voxelization, 2) retargetable volumetric scene reconstruction, 3) non-rigid point cloud registration and volumetric deformation, and 4) spatially varying geometry and appearance refinement. ASH and its example applications are open sourced in Open3D (\url{http://www.open3d.org}).
\end{abstract}

}
\maketitle

\IEEEdisplaynontitleabstractindextext
\IEEEpeerreviewmaketitle

\IEEEraisesectionheading{\section{Introduction}\label{sec:intro}}
\input{tex/introduction.tex}

\section{Related Work}\label{sec:related}
\input{tex/related-work.tex}

\section{Overview}\label{sec:overview}
\input{tex/overview.tex}

\section{The ASH Framework}\label{sec:architecture}
\input{tex/parallelhashing.tex}

\section{Experiments}\label{sec:experiment}
\input{tex/experiments.tex}

\section{Applications}\label{sec:applications}
\input{tex/application.tex}

\section{Conclusion and Future Work}\label{sec:conclusion}
\input{tex/conclusion.tex}

\bibliographystyle{IEEEtran}
\bibliography{paper}
\begin{IEEEbiography}[{\includegraphics[width=1in,height=1.25in,clip,keepaspectratio]{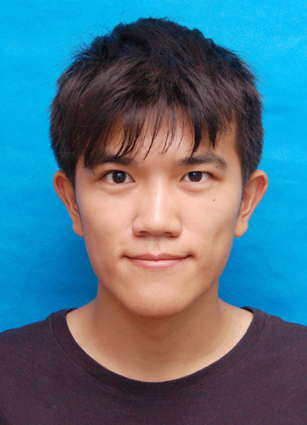}}]{Wei Dong}
  is a PhD student at Carnegie Mellon University since 2018, advised by Michael Kaess. Prior to PhD, he received his bachelor's degree (2015) and master's degree (2018) in Computer Science from Peking University, advised by Hongbin Zha.
\end{IEEEbiography}
\begin{IEEEbiography}[{\includegraphics[width=1in,height=1.25in,clip,keepaspectratio]{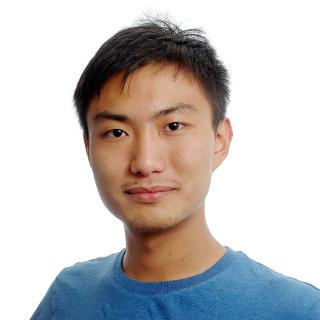}}]{Yixing Lao}
  is a Research Engineer at Intel Intelligent Systems Lab. He received his master's degree in Computer Science from University of California, San Diego in 2016, and bachelor's degree in Computer Engineering from the University of Hong Kong in 2013.
\end{IEEEbiography}
\begin{IEEEbiography}[{\includegraphics[width=1in,height=1.25in,clip,keepaspectratio]{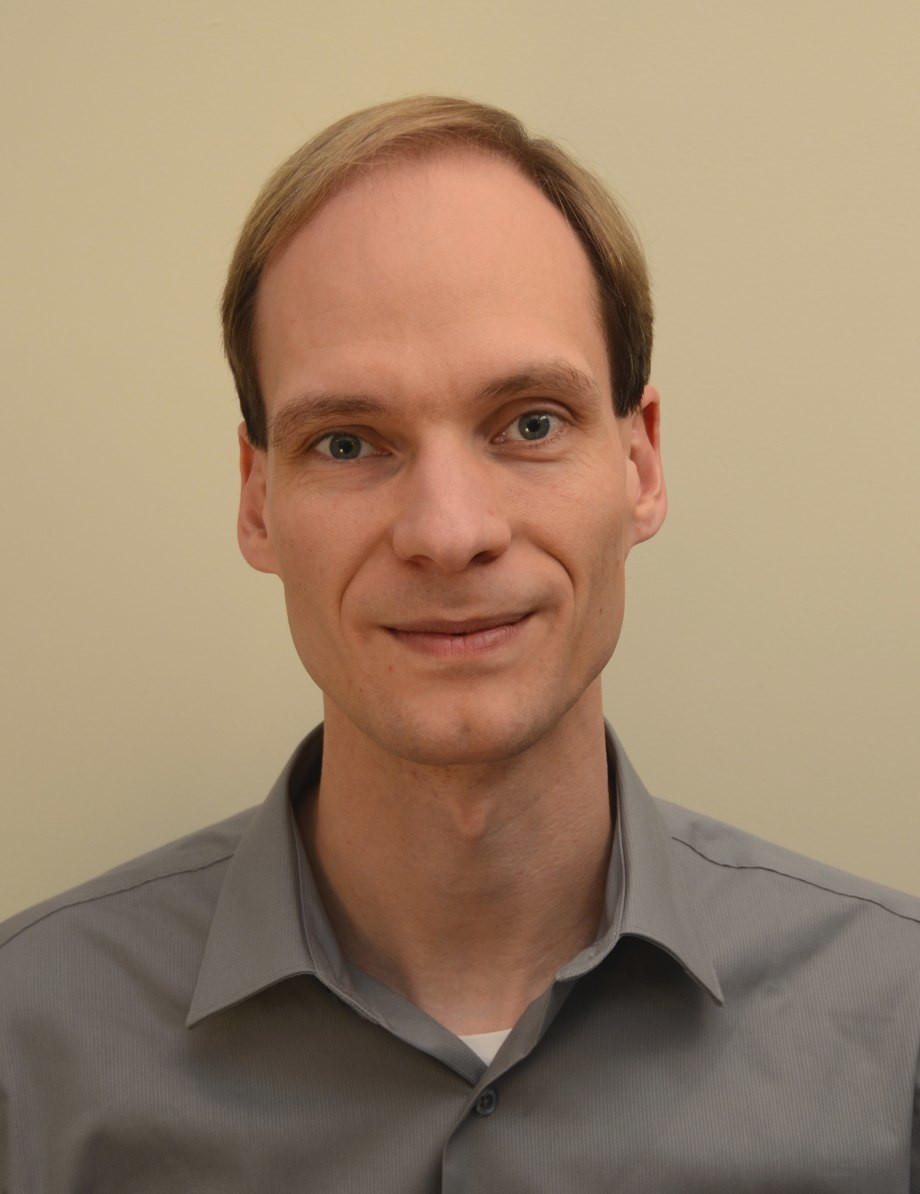}}]{Michael Kaess}
  is an Associate Professor in the Robotics Institute at Carnegie Mellon University and the Director of the Robot Perception Lab. He was a Research Scientist and a Postdoctoral Associate at the Massachusetts Institute of Technology. In 2008, he received the PhD degree in Computer Science from the Georgia Institute of Technology. He was Associate Editor for IEEE Transaction on Robotics and is currently Associate Editor for IEEE Robotics and Automation Letters. He is a Senior Member of IEEE.
\end{IEEEbiography}
\begin{IEEEbiography}[{\includegraphics[width=1in,height=1.25in,clip,keepaspectratio]{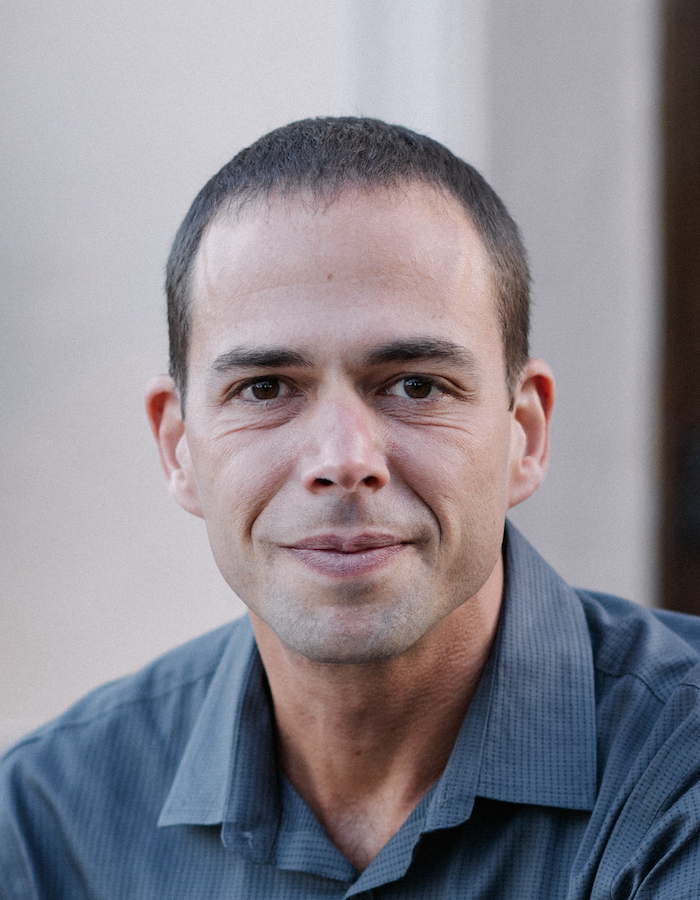}}]{Vladlen Koltun}
is a Distinguished Scientist at Apple. Previously he was the Chief Scientist for Intelligent Systems at Intel, where he built an international research lab, based on four continents, that produced high-impact results in robotics, computer vision, image synthesis, machine learning, and other areas. He has mentored more than 50 PhD students, postdocs, research scientists, and PhD student interns, many of whom are now successful research leaders.
\end{IEEEbiography}
\vfill
\end{document}

%% file: vcl-shortcuts.tex
\usepackage{epsfig}
\usepackage{graphicx}
\usepackage{float}
\usepackage{wrapfig}
\usepackage{amsmath,amssymb,amsthm}
\usepackage{algorithm,algorithmicx,algpseudocode}
\usepackage{bm,xspace}
\usepackage{comment}
\usepackage{verbatim}
\usepackage{multirow}
\usepackage{balance}
\usepackage{url}
\usepackage{booktabs}
\usepackage{etoolbox,siunitx}
\usepackage{calc}
\usepackage{pifont,hologo}
\usepackage[usenames, dvipsnames]{color}
\usepackage{nicefrac}
\usepackage{listings}
\usepackage{xcolor}
\usepackage{framed}
\usepackage{enumitem}

\definecolor{codegreen}{rgb}{0,0.6,0}
\definecolor{codegray}{rgb}{0.5,0.5,0.5}
\definecolor{codepurple}{rgb}{0.58,0,0.82}
\definecolor{backcolor}{rgb}{0.89,0.99,0.97}
\lstdefinestyle{mystyle}{
    backgroundcolor=\color{backcolor},
    commentstyle=\color{codegreen},
    keywordstyle=\color{magenta},
    numberstyle=\tiny\color{codegray},
    basicstyle=\ttfamily\footnotesize,
    keepspaces=true,
    numbers=left,
    numbersep=3pt,
    belowcaptionskip=5em,
    belowskip=3em,
    numbers=left,
    xleftmargin=0.8em,
    columns=fullflexible,
}
\newcommand{\ra}[1]{\renewcommand{\arraystretch}{#1}}

\usepackage[super]{nth}
\usepackage{nicefrac}
\robustify\bfseries

\def\aa{\mathbf{a}}

\def\cc{\mathbf{c}}
\def\dd{\mathbf{d}}

\def\hh{\mathbf{h}}
\def\ii{\mathbf{i}}

\def\kk{\mathbf{k}}

\def\nn{\mathbf{n}}

\def\pp{\mathbf{p}}
\def\qq{\mathbf{q}}

\def\tt{\mathbf{t}}
\def\uu{\mathbf{u}}
\def\vv{\mathbf{v}}

\def\xx{\mathbf{x}}
\def\yy{\mathbf{y}}

\def\CC{\mathbf{C}}

\def\NN{\mathbf{N}}

\def\RR{\mathbf{R}}

\def\TT{\mathbf{T}}

\def\iI{\mathcal{I}}

\def\kK{\mathcal{K}}

\def\mM{\mathcal{M}}

\def\qQ{\mathcal{Q}}

\def\vV{\mathcal{V}}

\def\xX{\mathcal{X}}
\def\yY{\mathcal{Y}}

\def\btheta{{\bm\theta}}

\DeclareMathOperator*{\argmin}{arg\,min}

\newcommand{\YesV}{\ding{51}}%
\newcommand{\NoX}{\ding{55}}%

\DeclareMathSymbol{@}{\mathord}{letters}{"3B}

\def\latex/{\LaTeX}
\def\bibtex/{\hologo{BibTeX}}

\def\eg{\textit{e.g.}}
\def\ie{\textit{i.e.}}

\newcommand\normaltt[1]{\normalfont\texttt{#1}}

\newtheorem{setup}{Setup}

%% file: tex/introduction.tex
\IEEEPARstart{3}{D} space is only one dimension higher than the 2D image space, yet the additional dimension introduces an unpredictable multiplier to the computational and storage cost. This increased dimension makes 3D perception tasks such as geometric reconstruction and appearance refinement challenging to implement.
To reduce complexity, one compromise is to reuse dense data structures and constrain the 3D space by bounding the region of interest, \eg, adopting a 3D array in a bounded space~\cite{newcombe2011kinectfusion}. While this simple approach succeeds at the scale of objects, it cannot meet the demand of room or city scale perception, which is necessary for virtual tour, telepresence, and autonomous driving.

Since the 3D space is generally a collection of 2D surface manifolds, its sparsity can be exploited by \emph{partitioning} to reduce computational cost. The general idea is to split the large 3D space into smaller regions and only proceed with the non-empty ones.
There is a plethora of well-established data structures for 3D space partitioning. Examples include trees (Octree~\cite{meagher1982octree}, KD-tree~\cite{bentley1975kdtree}) and hash maps (spatial hashing~\cite{niessner2013real}). While trees are able to adaptively achieve high precision, they 1) require an initial bounding volume and 2) usually take unbalanced traversal time for a batch of spatial queries and hence 3) are less friendly to batched operations.
On the other hand, spatial hashing coupled with a plain array structure is more scalable and parallelizable for reconstruction tasks.

In classical dense SLAM pipelines~\cite{niessner2013real,prisacariu2017infinitam}, spatial hashing is used to map 3D coordinates to internal pointers that are only accessible in GPU device code. A pre-allocated memory pool allows dynamic growing hash maps, but adaptive rehashing is generally unavailable. Recent feature-grid encoders~\cite{mueller2022instant} apply spatial hashing to map 3D coordinates to features stored at grid points in a \emph{static} bounded region. While differentiable spatial query is supported, \emph{collisions} are not resolved, limiting its usage to stochastic feature optimization where incorrect key-value mappings are tolerated.
A general, user-friendly, collision-free hash map is missing for efficient spatial perception at scale.

The reason for this absence is understandable. A parallel hash map on GPU has to resolve collisions and thread conflicts and preferably organize an optimized memory manager, none of which is trivial to implement. Previous studies have attempted to tackle the problem in one or more aspects, driven by their selected downstream applications. Furthermore, most of the popular parallel GPU hash maps are implemented in C++/CUDA and only expose low-level interfaces. As a result, customized extensions must start from low-level programming. While these designs usually guarantee performance under certain circumstances~\cite{alcantara2009real,ashkiani2018slab, niessner2013real, prisacariu2017infinitam}, as of today, they leave a gap from the standpoint of the research community, which prefers to use off-the-shelf libraries for fast prototyping with a high-level scripting language using tensors and automatic differentiation. Our motivation is to bridge this gap to enable researchers to develop sophisticated 3D perception routines with less effort and drive the community towards large-scale 3D perception.

To this end, we design a modern hash map framework with the following major contributions:
\begin{enumerate}[leftmargin=*]
    \item a user-friendly \emph{dynamic}, \emph{generic}\footnote{A dynamic hash map supports insertion and deletion after hash map construction. A generic hash map supports arbitrary dimensional keys and values in various data types.}, and \emph{collision-free} hash map interface that enables tensor I/O, \emph{advanced indexing}, and \emph{in-place automatic differentiation} when bridged to autodiff engines such as PyTorch;
    \item an index-first adaptor that supports various state-of-the-art parallel GPU hash map backends and accelerates hash map operations with an improved structure-of-array (SoA) data layout;
    \item a number of downstream applications that achieve higher performance compared to state-of-the-art implementations with fewer LoC.
\end{enumerate}
Experiments show that ASH achieves better performance with fewer LoC on both synthetic and real-world tasks.

%% file: tex/related-work.tex
\subsection{Parallel Hash Map}\label{subsec:parallel-hash-map}
The hash map is a data structure that seeks to map \textit{sparse keys} (\eg~unbounded indices, strings, coordinates) from the set $\kK$ to values from the set $\vV$ with amortized $O(1)$ access. It has a hash function $h: \kK \to \iI_n, k \mapsto h(k)$ that maps the key to the index set $\iI_n = \{0, 1, \dotsc, n-1\}$ for indexing (or addressing) that is viable on a computer.

Ideally, with a perfect injective hash function $h$, a hash map can be implemented by $H: \kK \to \vV, k \mapsto \vv^A\big[h(k)\big]$, where $\vv^A$ is an array of objects of type $\vV$ and $[\cdot]$ is the trivial array element accessor. However, in practice, it is intractable to find an injective map given a sparse key distribution in $\kK$ and a constrained index set $\iI_n$ of size $n$ due to the computational budget. Therefore, modifications are required to resolve inevitable \emph{collisions}, where $i = h(k_1) = h(k_2), k_1 \neq k_2$.
There are two classes of techniques for collision resolution, \emph{open addressing} and \emph{separate chaining}. Open addressing searches for another candidate $j \neq i, j \in \iI_n$ via a \emph{probing} algorithm until an empty address is found. The simplest probing, linear probing~\cite{karnagel2015optimizing}, computes $j = (h(k) + t) \mod{n}$ starting from $i=h(k)$, where $t$ is the number of attempts. Separate chaining, on the other hand, maintains multiple entries per mapped index where a linked list is grown at $i$ if $i = h(k_1) = h(k_2), k_1 \neq k_2$.

While hash map implementations are widely available for CPU, their GPU counterparts have only emerged in the recent decade. Most GPU hash maps use open addressing~\cite{cudf, alcantara2009real, garcia2011coherent, junger2020warpcore}, mainly due to simplicity in implementation and capability of handling highly concurrent operations. CUDPP~\cite{alcantara2009real} utilizes Cuckoo Hashing~\cite{pagh2004cuckoo}, while CoherentHash~\cite{garcia2011coherent} adopts Robin Hood Hashing~\cite{celis1985robin} -- both involving advanced probing design. Although being performant when $\kK, \vV$ are limited to integer sets, these variations cannot be generalized to spatial hashing and only allow static input. Recently, WarpCore~\cite{junger2020warpcore} proposes to support non-integer $\vV$ and dynamic insertion, but the key domain is still limited to at most 64 bits.

There are also a few separate chaining implementations on GPU involving device-side linked lists.
SlabHash~\cite{ashkiani2018slab} builds a linked list with a 128-bit \emph{Slab} as the minimal unit, optimized for Single Instruction Multiple Threads (SIMT) warp operations. Although SlabHash allows dynamic insertions, similar to the aforementioned GPU hash maps, only integer $\kK, \vV$ are supported. stdgpu~\cite{stotko2019stdgpu} follows the conventional C++ Standard Library {\ttfamily std::unordered\_map} and builds supporting vectors, bitset lock guards, and linked lists from scratch, resulting in a generic, dynamic hash map. With these rich functionalities, however, stdgpu is not optimized for large value sets. In addition, due to its low-level templated design, users have to write device code for simple tasks.

We refer the readers to a comprehensive review of GPU hash maps~\cite{lessley2019data}.

\begin{table}
  \caption{Comparison of existing parallel GPU hash maps. ASH preserves the dynamic, generic, and atomic properties, and is extendable to the non-templated high-level Python interfaces.}
  \newcolumntype{Z}{S[table-format=2.1,table-auto-round]}
  \centering
  \ra{1.05}
  \small
  \resizebox{0.95\linewidth}{!}{
  \begin{tabular}{@{}lcccc@{}}
    \toprule
                                             & {Dynamic} & {Generic} & {Collision-free} & {Python} \\
    \midrule
    SlabHash~\cite{ashkiani2018slab}         & \YesV     & \NoX      & \YesV    & \YesV    \\
    CUDPP~\cite{alcantara2009real}           & \NoX      & \NoX      & \YesV    & \NoX     \\
    cuDF~\cite{cudf}                         & \NoX      & \NoX      & \YesV    & \YesV    \\
    WarpCore~\cite{junger2020warpcore}       & \YesV     & \NoX      & \YesV    & \NoX     \\
    stdgpu~\cite{stotko2019stdgpu}           & \YesV     & \YesV     & \YesV    & \NoX     \\
    \midrule
    InfiniTAM~\cite{prisacariu2017infinitam} & \YesV     & \NoX      & \YesV    & \NoX     \\
    VoxelHashing~\cite{niessner2013real}     & \YesV     & \NoX      & \NoX     & \NoX     \\
    GPURobust~\cite{dong2019gpu}             & \YesV     & \YesV     & \NoX     & \NoX     \\
    Instant-NGP~\cite{mueller2022instant}    & \NoX      & \YesV     & \NoX     & \YesV     \\
    \midrule
    ASH                                      & \YesV     & \YesV     & \YesV    & \YesV    \\
    \bottomrule
  \end{tabular}
  }
  \label{tab:comparison}
  \vspace{-2mm}
\end{table}

\subsection{Space Partitioning Structures}\label{subsec:3d-representation}
3D data is not as simple to organize as 2D images. While a 2D image can be stored in a dense matrix, exploiting sparsity in 3D data is paramount due to the limits in computer memory of the current day.

The most widely used data structures for 3D indexing are arguably trees.
A KD-tree~\cite{bentley1975kdtree} recursively sorts k-dimensional data along a selected axis and partitions data at the median point. By nature, a KD-Tree is designed for neighbor search. In 3D, it is mainly used to organize 3D \emph{points} and their \emph{features}. Examples include normal estimation and nearest neighbor association in Iterative Closest Points (ICP)~\cite{rusinkiewicz2001efficient,zhou2018open3d} and 3D feature association in global registration~\cite{rusu2009fast,Zhou2016}. GPU adaptations exist for KD-trees\cite{rusu20113d, zhou2008kdtree}, but are not suitable for incrementally changing scenes, as they are usually constructed once and queried repeatedly.

Bounding volume hierarchy (BVH) is another hierarchical representation that organizes primitives such as \emph{objects} and \emph{triangles} in 3D.
There are various GPU adaptations~\cite{gunther2007realtime, lauterbach2009fast, wald2007ray} mostly targeted at ray tracing and dynamic collision detection. While a parallel construction is possible and deformation of the nodes is allowed, the tree structure typically remains unchanged, assuming a fixed layout.

While KD-trees and BVH split the space \emph{unevenly} by data distribution, an Octree~\cite{meagher1982octree}, on the other hand, recursively partitions the 3D space \emph{evenly} into 8 subvolumes according to space occupation states. It has been widely used in adaptive 3D mapping~\cite{hornung2013octomap, museth2013openvdb} for robot navigation. There have been parallel implementations on GPU, from optimized data structures~\cite{hoetzlein2016gvdb} to domain-specific languages~\cite{hu2019taichi}. However, these works generally focus on physics simulation within a bounded region of interest where the spatial partition is predefined. While parallel incremental division~\cite{zeng2013octree} is possible, an initial bounding region is still required, and the trees are not guaranteed to be balanced.

Spatial hashing is another variation of spatial management with $O(1)$ access time depending on hash maps. Bundled with small dense 3D arrays, it has been widely used in real-time volumetric scene reconstruction within \emph{unbounded} region of interest. A handful of CPU implementations have achieved real-time performance~\cite{klingensmith2015chisel,han2018flashfusion} at the expense of resolution. Similarly, GPU implementations \cite{niessner2013real, prisacariu2017infinitam,dong2019gpu, dong2018psdf} reach high frame rates using GPU-based spatial hashing. However, all of the studies depend on ad hoc GPU hash maps exclusive to these specific systems. Concurrent race conditions have not been fully resolved in several implementations~\cite{niessner2013real, dong2019gpu}, where volumes can be randomly under-allocated. Recent neural feature grids~\cite{mueller2022instant} apply spatial hashing in a \emph{bounded} volume to query voxelized feature embeddings. These approaches use simplified hashing designs that are not collision-free, and thus are only compatible with stochastic optimization that tolerates noise from an incorrect query.

\begin{figure*}[ht]
  \includegraphics[width=\linewidth]{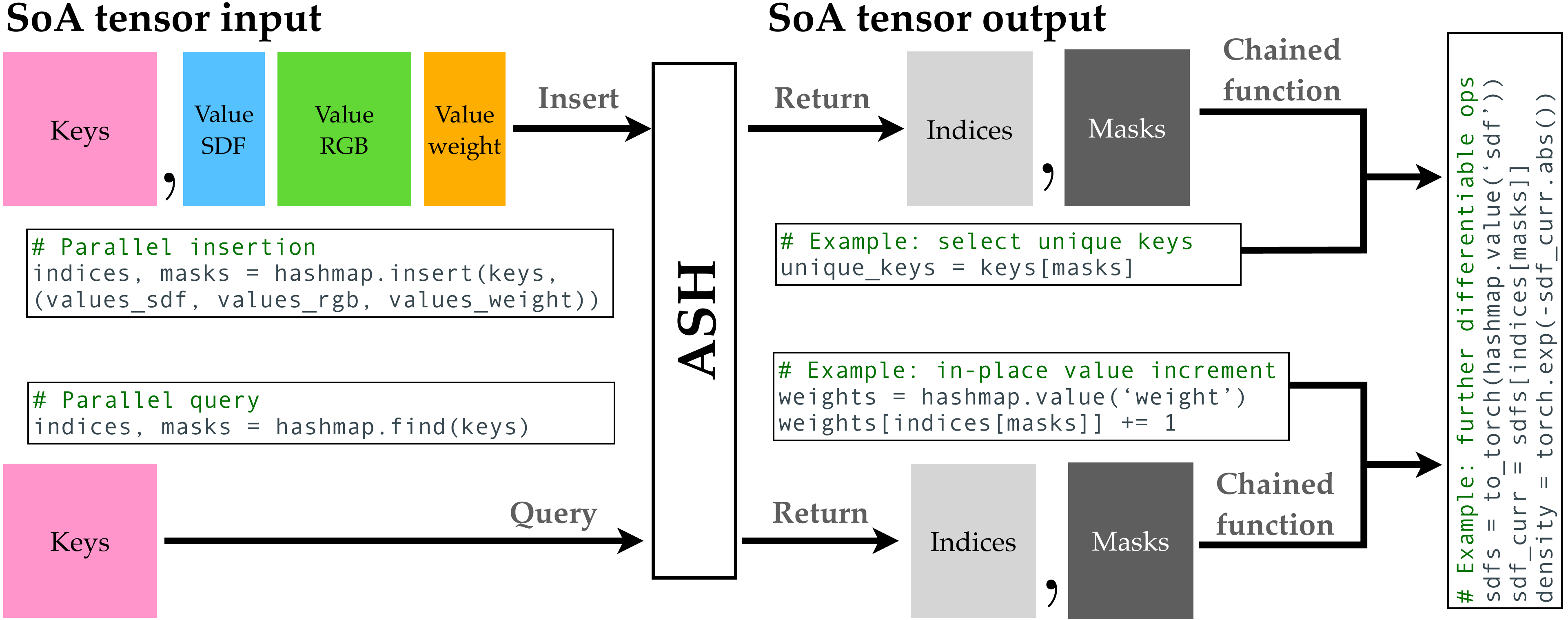}
  \caption{The interface illustration of ASH. Left: ASH takes a key and/or a list of value tensors as input, with a one-liner interface. For \emph{insertion}, tensors are organized in SoA.
    Right: buffer indices and masks are returned upon insertion and query, organized together in SoA. Chained functions can be easily applied to hashed data by indexing ASH buffers with indices and masks. Examples include selecting unique keys through insertion, and applying in-place value increment through query. Differentiable ops can be applied in downstream applications.}
  \label{fig:hash-map-interface-ash}
\end{figure*}

\subsection{Spatially Varying 3D Representations}
A truncated signed distance function (TSDF)~\cite{curless1996volumetric} is an implicit representation of surfaces, recording point-wise distance to the nearest surface point. It is frequently used for dense scene reconstruction with noisy input. The distribution of surfaces is generally spatially varying and therefore, a proper parameterization is often necessary, either in a discrete~\cite{niessner2013real} or neural~\cite{chabra2020deep, mueller2022instant} form.
Non-rigid deformation methods~\cite{zhou2014slac, newcombe2015dynamicfusion} seek to embed point clouds in a deformable grid, where each point is anchored to and deformed by neighbor grids. They are mainly used for animation or non-rigid distortion calibration.
Similar to a deformation grid, complex lighting for rendering can be approximated by spatially-varying spherical harmonics (SVSH)~\cite{maier2017intrinsic3d} placed at a sparse grid. These grids are natural applications of spatial hashing.
A comprehensive review of spatially varying representations for real-world scene reconstruction is available~\cite{zollhofer2018state}.

While ad hoc implementations have been introduced for these representations either on CPU or GPU, ASH provides a device-agnostic interface requiring less code written and providing better performance.
Table~\ref{tab:comparison} compares various aspects of existing GPU hash maps, either as a standalone data structure (Section~\ref{subsec:parallel-hash-map}) or embedded in an application (Section~\ref{subsec:3d-representation}). To the best of our knowledge, ASH is the first implementation that simultaneously supports dynamic insertion, ensures correctness via atomic collision-free operations, allows generic keys, and has a modern tensor interface and Python binding for better usability.

%% file: tex/overview.tex
Before plunging into the details, we first provide a high-level overview of our framework in Fig.~\ref{fig:hash-map-interface-ash}.

Conventional parallel hash maps reorganize the structure of arrays (SoA) input, \ie, the separated key array and value array, into an array of structures (AoS) where keys and values are \emph{paired}, inserted, and stored. Therefore, array of \emph{pointers to pair structures} ({\ttfamily std::pair} in C++, {\ttfamily thrust::pair} in CUDA, and {\ttfamily tuple} in Python) are returned upon query.
Consequently, the operations from insertion and query to in-place value increment require users to write device code and visit AoS at the low-level pointers.

In contrast, ASH sticks to SoA. Fig.~\ref{fig:hash-map-interface-ash} shows the workflow of ASH. Instead of \emph{pointers to pairs}, ASH returns \emph{indices} and \emph{masks} arrays that can be directly consumed by tensor libraries such as PyTorch~\cite{paszke2019torch} (without memory copy) and NumPy~\cite{harris2020array} (with GPU to host memory copy).
As a result, post-processing functions such as duplicate key removal and in-place modification can be chained with insertion and query in ASH via advanced indexing without writing any device code.
As a general and device-agnostic interface for parallel hash maps, our framework is built upon switchable backends with details hidden from the user. Currently, \emph{separate chain} backends are supported, including the generic stdgpu~\cite{stotko2019stdgpu} backend, and the extended integer-only SlabHash~\cite{ashkiani2018slab} backend for arbitrary key-value data types. TBB's concurrent hash map~\cite{kukanov2007foundations} powers the CPU counterpart with the identical interface to GPU.

In this paper, we use calligraphic letters to represent sets, and normal lower-case letters for their elements. Normal upper-case letters denote functions. Bold lower-case letters denote vectors of elements or arrays in the programmer's perspective. Bold upper-case letters are for matrices.
For instance, in a hash map, we are interested in key elements $k \in \kK$ and their vectorized processing, \eg~query $Q(\kk)$. Specifically, we use $\iI_n$ to denote a set of indices $\{0, 1, \dotsc, n-1\}$, and $\Theta$ as the boolean selection $\{0, 1\}$. Given an arbitrary vector $\xx$, we denote $\xx(\ii)$ and $\xx(\btheta)$ as indexing and selection functions applied to $\xx$ when $\forall i \in \ii, i \in \iI_n$ and $\forall \theta \in \btheta, \theta \in \Theta$. We use $\langle \kK, \vV \rangle$ as the key and value sets for a hash map. $h: \kK \to \iI_n$ is the internal hash function that converts a key to an index. $H: \kK \to \vV$ is the general hash map enclosing $h$.

%% file: tex/parallelhashing.tex
\subsection{Classical Hashing}
In a hash map $\langle \kK, \vV \rangle$, since the hash function $h$ cannot be perfect as discussed in Section~\ref{sec:related}, we have to store keys to verify if collisions happen ($h(k_1) = h(k_2)$ but $k_1 \neq k_2, k_1, k_2 \in \kK$).

\begin{figure}[ht]
    \centering
    \includegraphics[width=\columnwidth]{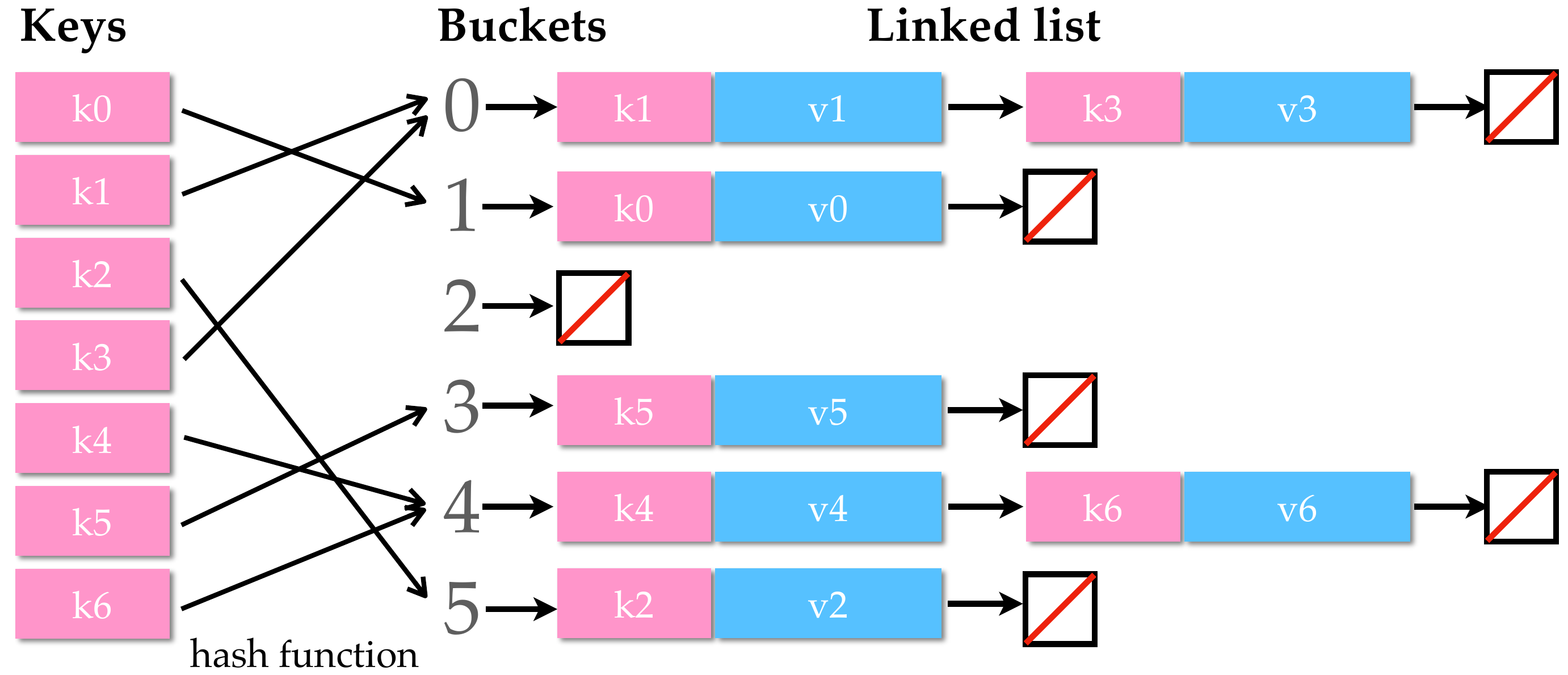}
    \caption{Illustration of a classical hash map using separate chaining. Keys (left) are put into corresponding buckets (middle) obtained by the hash function $h$. A linked list (right) is constructed per bucket to store key-value pairs within the same bucket but with unequal keys.}
    \label{fig:hash-map-classical}
\end{figure}

In separate chaining, to resolve hash collisions, the bucket-linked list architecture is used.
With $n$ initial \emph{buckets}, we construct the hash function $h: \kK \to \iI_n$ where $\iI_n$ is defined in Section~\ref{sec:related}.
As shown in Fig.~\ref{fig:hash-map-classical}, keys with the same hashed index $i = h(k) \in \iI_n$ are first aggregated in the $i$-th bucket, where a linked list grows adaptively to accommodate different keys. A conventional hash map stores key-value pairs as the storage units. Consequently, two keys $k_1, k_2$ can be distinguished by checking $h(k_1) = h(k_2)$ and $k_1 = k_2$ from the pair in order, and manipulation of the keys and values can be achieved by iterating over such pairs.

With this formulation, assuming a subset $\xX \subset \kK$ has been inserted into the hash map with associated values $\yY \subset \vV$, a query function can be described as
\begin{align}
    Q_{\kK, \vV}: \kK &\to \kK \times \vV \nonumber\\
  k &\mapsto \langle k, v\rangle,~\forall k \in \xX,\label{eq:query-elemwise}
\end{align}
where $\langle k, v \rangle$ forms a concrete pair stored in the hash map. This format is common in implementations, \eg~in C++ ({\ttfamily std::unordered\_map}) and Python ({\ttfamily dict}).

\subsection{Function Chaining and Parallel Hashing}

The element-wise operation in Eq.~\ref{eq:query-elemwise} can be extended to vectors via parallel device kernels. However, interpretation of the returned iterators of pairs is still required at the low level. In other words, although the parallel version can be implemented efficiently, results are still packed in an AoS instead of SoA:
\begin{align}
    Q_{\kK, \vV}(\kk) = \mathbf{array}\{\langle k, v \rangle\}.
\end{align}

This forms a barrier when the parallel query is located in a chain of functions. For instance, to apply any function $G$ (\eg, geometry transformation) over the result of a query, the low-level function {\ttfamily second} that selects the value element from a pair $\langle k, v \rangle$ must be provided to dereference the low-level structures and manipulate the keys and values \emph{in-place}. In other words, we have to implement a non-trivial $\tilde{G}$:
\begin{align}
\tilde{G}(\kk) = (G \circ \texttt{second}\circ Q_{\kK, \vV}) (\kk),
\end{align}
to force the conversion from AoS to SoA and chain a high-level function $G$ with $Q_{\kK, \vV}$
\footnote{This can be achieved simply by returning a \emph{copy} of values, but it is not feasible, especially when dealing with large-scale data, \eg~hierarchical voxel grids.}.
This could be tedious when prototyping geometry perception that requires hash map structures since off-the-shelf operations have to be reimplemented in device code.

We reformulate this problem by introducing two affiliate arrays, $\kk^B$ and $\vv^B$ (note with a superscript $B$ for buffering, they are not the input $\kk, \vv$) of capacity $c \ge n$, where $n$ is the number of buckets.
These arrays are designed for explicit storage of keys and values, respectively, and serve as buffers to support natural SoA. They are exposed to users for direct access and in-place modification. Now the query function can be rewritten as
\begin{align}
    Q_{\kK, \vV}: \kK &\to \iI_c, k \mapsto i, \nonumber \\
    s.t.~~&\kk^B(i) = k,~\vv^B(i) = v,~\forall \langle k, v\rangle \in \langle \xX, \yY \rangle
\end{align}
and this version is ready for parallelization.
At this stage, to combine $G$ and $Q_{\kK, \vV}$, we can chain
$G \circ \vv^B \circ Q$ to manipulate values:
\begin{align}
    G(\kk) = G \bigg(\vv^B(Q_{\kK, \vV}(\kk))\bigg),
\end{align}
which retains convenient properties such as array vectorization and advanced indexing.

When the input set $\tilde{\xX} \not\subset \xX$ is not fully stored in the hash map, our formulation maintains its effectiveness by a simple masked extension:
\begin{align}
    Q_{\kK, \vV}^\iI: \kK &\to \iI_c, ~Q_{\kK, \vV}^{\Theta} \to \{0, 1\}, \nonumber \\
    Q_{\kK, \vV}^\iI(\tilde{k}) &= i, ~\qQ_{\kK, \vV}^{\Theta} = \theta, \nonumber \\
    ~~s.t.~~\theta &= 1;~\kk^B(i) = \tilde{k},~\vv^B(i) = \tilde{v}, ~\mathrm{if}~\tilde{k} \in \xX, \\
    \theta &= 0; i = \mathrm{undefined}, ~\mathrm{otherwise}, \nonumber
\end{align}
which is also ready for parallelization. Now the chaining of functions is given by
\begin{align}
    G(\kk) &= G \bigg(\vv^B\Big(\ii (\btheta) \Big) \bigg),\\
    \ii &= Q_{\kK, \vV}^{\iI}(\kk),\btheta = Q_{\kK, \vV}^{\Theta}(\kk),
\end{align}
using advanced indexing with masks. We can also select valid queries with $\kk(\btheta)$ without visiting $\kk^B$. While our discussion was about the query function, the same applies to insertion.

In essence, by converting the pair-first AoS to an index-first SoA format with the help of array buffers, we can conveniently chain high-level functions over hash map query and insertion.
This simple change enables easy development on hash maps and unleashes their potential for fast prototyping and differentiable computation. However, the layout requires fundamental changes to the hash map data structure. With this in mind, we move on to illustrate how the ASH layer converts the AoS in native backends to our SoA layout.

\subsection{Generic Backends}\label{subsec:generic}
We start with converting stdgpu~\cite{stotko2019stdgpu}, a state-of-the-art generic GPU hash map as the backend of ASH. stdgpu follows the convention of its CPU counterpart {\ttfamily std::unordered\_map} by providing a templated interface. The underlying implementation is a classical bucket - linked list structure with locks to avoid race conditions on GPU. To exploit the power of a generic hash map without reinventing the wheel, we seek to reuse the operations over keys (\ie~lock-guarded bucket and linked list operations) and redirect the value mapping to our buffer $\vv^B$.

\begin{figure}[ht]
    \centering
    \includegraphics[width=\columnwidth]{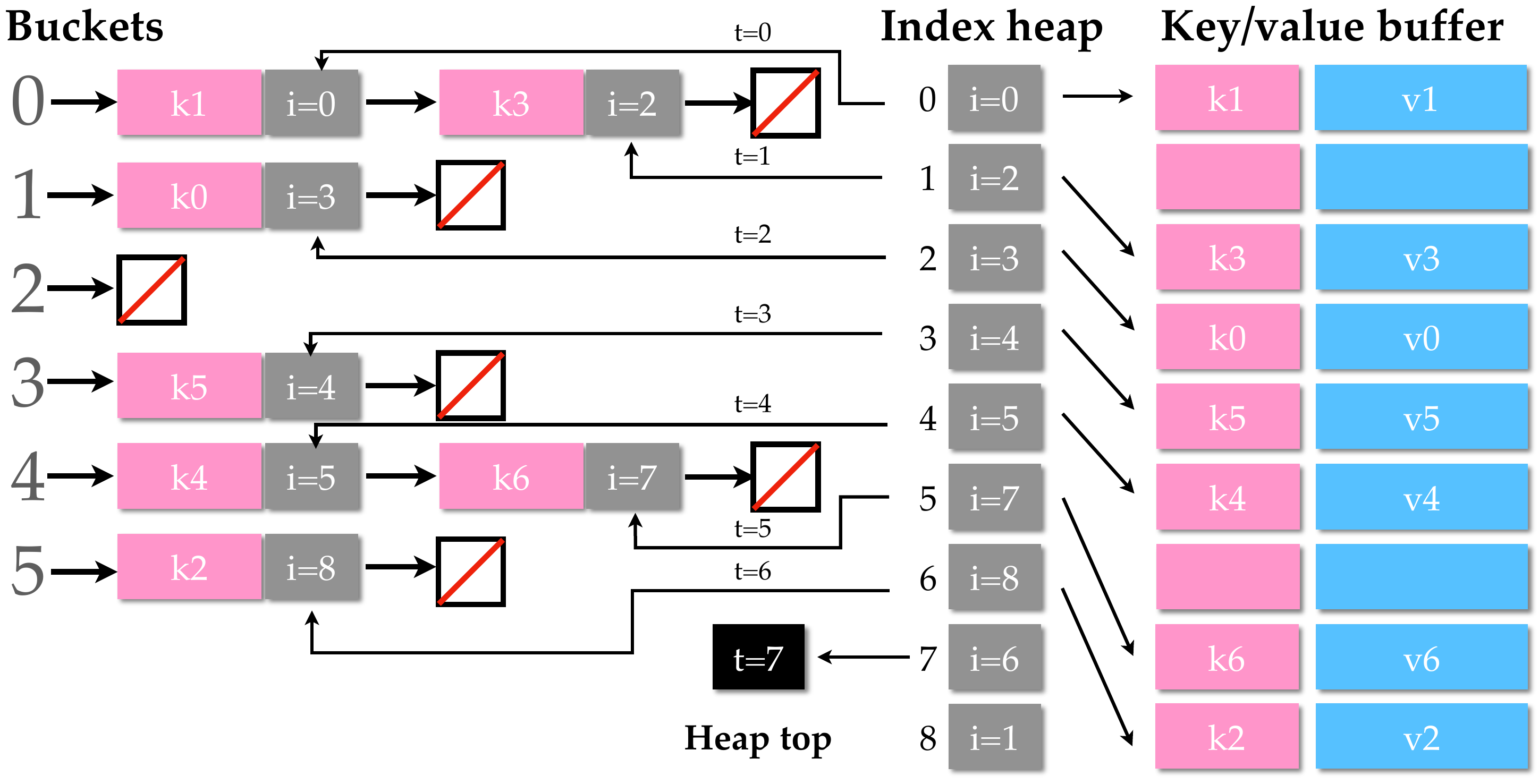}
    \caption{Illustration of a generic hash map bridged to tensors in ASH. \emph{Buffer indices} $i$ are dynamically provided by the available indices maintained in the index heap at the increasing heap top $t$ (middle), acting as the \emph{values} in the underlying hash map (left). It connects the hash map and the actual key values stored in the buffer (right) by accessing $i$. The key-bucket correspondences are the same as Fig.~\ref{fig:hash-map-classical}, omitted for simplicity.}\label{fig:hash-map-generic}
  \end{figure}

A dynamic GPU hash map requires dynamic allocation and freeing of keys and values in device kernels. With pre-allocated key buffer $\kk^B$ and value buffer $\vv^B$, we maintain an additional \emph{index heap} $\hh$, as shown in Fig.~\ref{fig:hash-map-generic}. The index heap stores \emph{buffer indices} $i$ pointing to the buffers $\kk^B, \vv^B$ as a map $P: \iI_c \to \iI_c$, where the \emph{heap top} $t$ maintains the currently available buffer index in $\hh[t]$. Heap top starts at $t = 0$, and is atomically increased at allocation and decreased at free.
With $\hh$ and the dynamically changing $t$, we instantiate a generic hash map with the templated value in stdgpu to be $\vV = \mathtt{Int32}$, where the values are \emph{buffer indices} $i$ stored in $\hh$ to access $\kk^B, \vv^B$ exposed to the user.

\subsubsection{Insertion}\label{subsec:generic-insertion}
The insertion of a $\langle k, v \rangle \in \langle \kK, \vV \rangle$ pair is now decoupled into two steps, with i) insertion of $\langle k, i\rangle \in \langle \kK, \iI_c \rangle$ into the hash map, where $i$ is the buffer index dynamically acquired from the heap top $\hh[t]$ and ii) insertion of $\kk^B(i):=k, \vv^B(i):=v$ into buffers.

A naive implementation will acquire a buffer index $i$ from $\hh$ on every insertion attempt and free it if the insertion fails because the key already exists. However, when running in parallel, {\ttfamily atomicAdd} and {\ttfamily atomicSub} may be conflicting among threads, leading to race conditions. A \emph{two-pass insertion} could resolve the issue: in the first pass, we allocate a batch of indices from $\hh$ determined by the input size, attempt insertions, and record results; in the second pass, we free the indices to $\hh$ from failed insertions.

We adopt a more efficient \emph{one-pass} \emph{lazy} insertion. We first attempt to insert $\langle k, -1\rangle$ with $-1$ as the dummy index into the backend and observe if it is successful. If not, nothing needs to be done. Otherwise, we capture the returned pointer to the pair, trigger an index $i$ allocation from $\hh$, and directly replace the dummy -1 with $i$. This significantly reduces the overhead when the key uniqueness is low (\ie, many duplicates exist in the keys to be inserted).

\subsubsection{Query}
The query operation is relatively simpler. We first look up the buffer index $i \in \iI_c$ given $k$ in the backend. If it is a success, we end up with $k = \kk^B(i)$, and the target $v = \vv^B(i)$ is accessible with $i$ by users.

\subsection{Non-generic Backends}\label{subsec:nongeneric}
While the generic GPU hash map has only recently been available, the research community in parallel computation has been focusing on more controlled setups where both $\kK$ and $\vV$ are limited to certain dimensions or data types. We seek to generalize this non-generic setup with our index heap and verify their performance in more real-world applications. In this section, we show how ASH can be used to generalize SlabHash~\cite{ashkiani2018slab}, a warp-oriented dynamic GPU hash map that only allows insertions and queries to {\ttfamily Int32} data type.

An extension to generic key types is non-trivial for SlabHash since its warp operations only apply to variables with limited word length. Our implementation extends the hash set variation of SlabHash, where only integers as keys are maintained in the backend.

\subsubsection{Generalization via Index Heap}
\begin{figure}[ht]
    \centering
    \includegraphics[width=\linewidth]{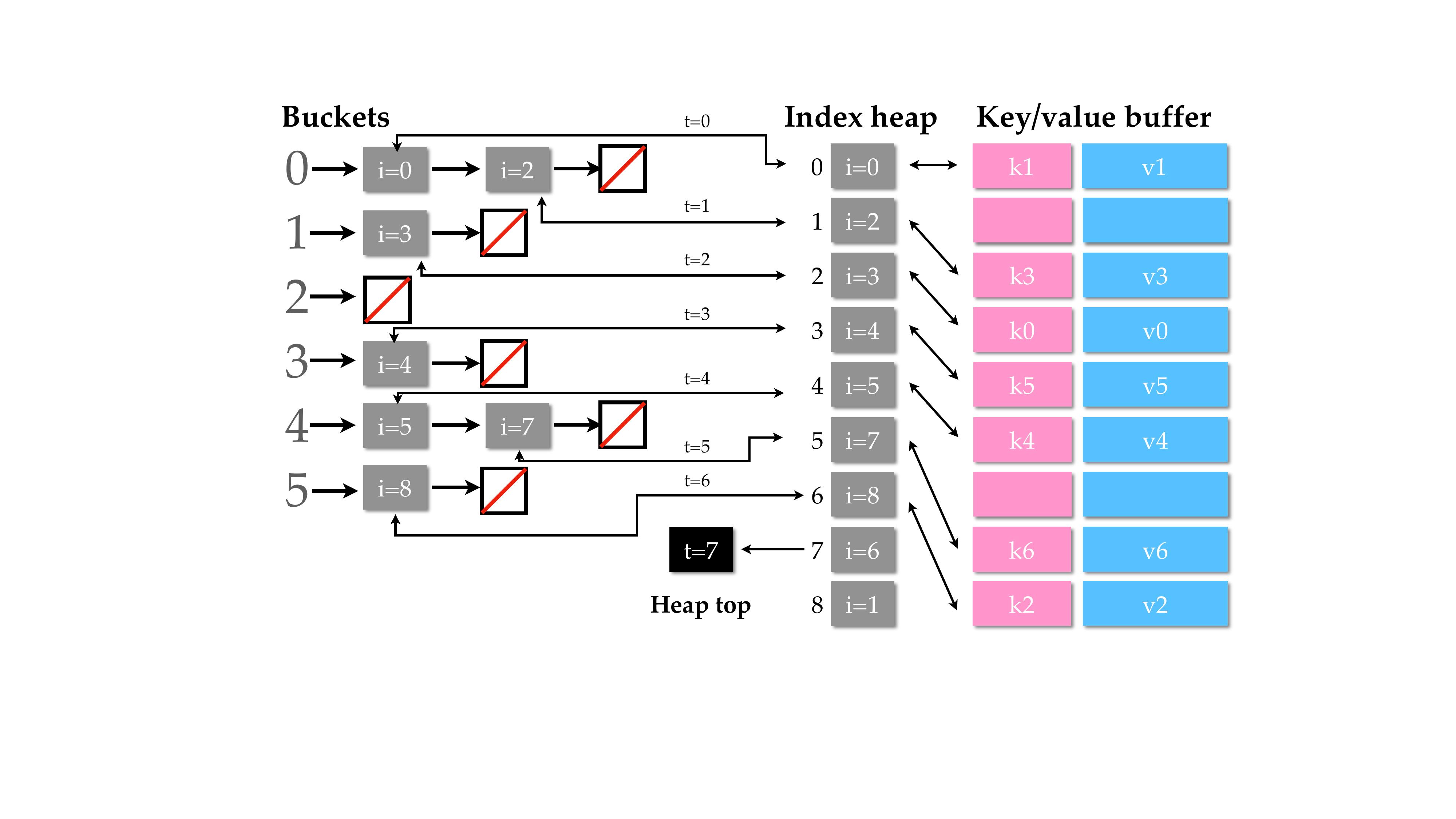}
    \caption{Illustration of a non-generic hash set enhanced by ASH. Integer \emph{buffer indices} $i$ allocated from the index heap (middle) are inserted as delegate \emph{keys} directly into the hash set (left), associated with \emph{actual keys} in the buffer (right) at $i$.
    The key-bucket correspondences are the same as Figs.~\ref{fig:hash-map-classical} and \ref{fig:hash-map-generic}, omitted for simplicity.}
    \label{fig:hash-map-nongeneric}
\end{figure}

The index heap $\hh$ is the core to generalizing the SlabHash backend. In brief, a generic key is represented by its associated buffer index $i$ in an integer-only hash set, allocated the same way as discussed in Section~\ref{subsec:generic-insertion}. As illustrated in Fig.~\ref{fig:hash-map-nongeneric}, all the insertions and queries are redirected from the buffer indices to actual keys and values via the index heap. However, the actual implementation involves more complicated changes in design.

Given a generic key $k$, we first locate the bucket $b = h(k) \in \iI_n$. Ideally, we can then allocate a buffer index $i$ at $\hh$'s top $t$ and insert it into the linked list at the bucket $b$ in the integer-only hash set. The accompanying key and value are put in $\kk^B(i), \vv^B(i)$. During query, we similarly first locate the bucket $b$ then search the key in the linked list by visiting $\kk^B$ via the stored index $i$.

\subsubsection{Multi-Pass Insertion}
Although query can be applied as mentioned above, lazy insertion mentioned in Section~\ref{subsec:generic-insertion} is problematic in this setup. The main reason is that while the race condition in inserting index $i$ does not occur in warp-oriented insertions, the copy of the \emph{actual} key $k \in \kK$ to $\kk^B$ requires global memory write.
They may not be synchronized among threads, as copying a multi-dimensional key takes several non-atomic instructions. As a result, the insertion of a key $k_2$ could be accidentally triggered when i) a duplicate $k_1 (= k_2)$'s index $i_1 \in \iI_c$ has been inserted but ii) whose actual key $k_1$ has only been partially copied to the buffer $\kk^B$. This would mistakenly result in $\kk^B(i_1) = k_1 \neq k_2$ followed by the unexpected insertion of $k_2$ when unsynchronized. In practice, with more than 1 million keys to be inserted in parallel, these kinds of conflicts happen with probability as low as $\le 0.1\%$.
To resolve conflicts, we split insertion into three passes:
\begin{itemize}[leftmargin=*]
    \item \emph{Pass 1:} batch insert \emph{all} keys $\kk$ to $\kk^B$ by directly copying all candidates via batch allocated corresponding indices $i$ from $\hh$;
    \item \emph{Pass 2:} perform parallel hashing with indices $i$ from pass 1. In this pass, keys are read-only in global buffers and hence do not face race conditions. Successful insertions are marked in a mask array.
    \item \emph{Pass 3:} batch insert values to $\vv^B$ with successful masks, and free the rest to $\hh$.
\end{itemize}
While there is overhead due to the multi-pass operation, it is still practical for a dynamic hash map. First, keys are relatively inexpensive to copy, especially for spatial coordinates, while the more expensive copying of values is done without redundancy.
Second, a dynamic hash map generally reserves sufficient memory for further growth so that the \emph{all} key insertion would not exceed the buffer capacity.

\subsection{Rehashing and Memory Management}
While buffers are represented as fixed-size arrays, growth of storage is needed to accommodate the accumulated input data, which can exceed the hash map's capacity, \eg~3D points from an RGB-D stream. This triggers rehashing, where we adopt the conventional $\times 2$ strategy to double the buffer size as common in the C++ Standard Library, collect all the active keys and values, and batch insert them into the enlarged buffer.

In dynamic insertions, there can be frequent free and allocation of small memory blobs that are adjacent and mergeable. In view of this, we implement another tree-structured global GPU manager similar to PyTorch~\cite{paszke2019torch}.

\subsection{Dispatch Routines}
To enable bindings to non-templated languages, \eg~Python, the tensor interface is non-templated so that it can take data types and shapes as arguments. In the context of spatial hashing, we support arbitrary dimensional keys by expanding the dispatcher macros in C++. Float types have undetermined precision behaviors on GPU. Therefore, a conversion to the integers given the desired precision is recommended to use the hash map.

We also additionally dispatch values by their element byte sizes into intrinsically supported vectors: {\ttfamily int, int2, int3}, and {\ttfamily int4}. This adaptation accelerates trivially copiable value objects such as {\ttfamily int3}, and supports non-trivially copiable value blocks (\eg~an $8^3$ array pointed to a {\ttfamily void} pointer).
This improves the insertion of large value chunks by a factor of 10 approximately.

\subsection{Multi-value Hash Map and Hash Set}
ASH supports multi-value hash maps that store values organized in SoA, as well as hash sets with only keys and no values.

\noindent \textbf{Multi-value hash maps}. Various applications in 3D processing require mapping coordinates to several properties. For instance, a 3D coordinate can be mapped to a normal, a color, and a label in a point cloud. While the mapped values can be packed as an array of structures (\ie, AoS) to fit a hash map, code complexity could increase since structure-level functions have to be implemented.
We generalize the hash map's functionality by extending the \emph{single} value buffer $\vv^B$ to an \emph{array of value buffers} $\{\vv^B_i\}$ and applying loops over properties per index during an insertion. This simple change supports the storage of complex value types in SoA that allows easy vectorized query and indexing.

\noindent \textbf{Hash set}. A hash set, on the other hand, is a simplified hash map -- an unordered set that stores unique keys. It is generally useful in maintaining a set by rejecting duplicates, such as in point cloud voxelization.
By removing $\vv^B$ and ignoring value insertion, a hash map becomes a hash set.

%% file: tex/experiments.tex
\begin{figure*}[ht]
    \centering
    \includegraphics[width=\linewidth]{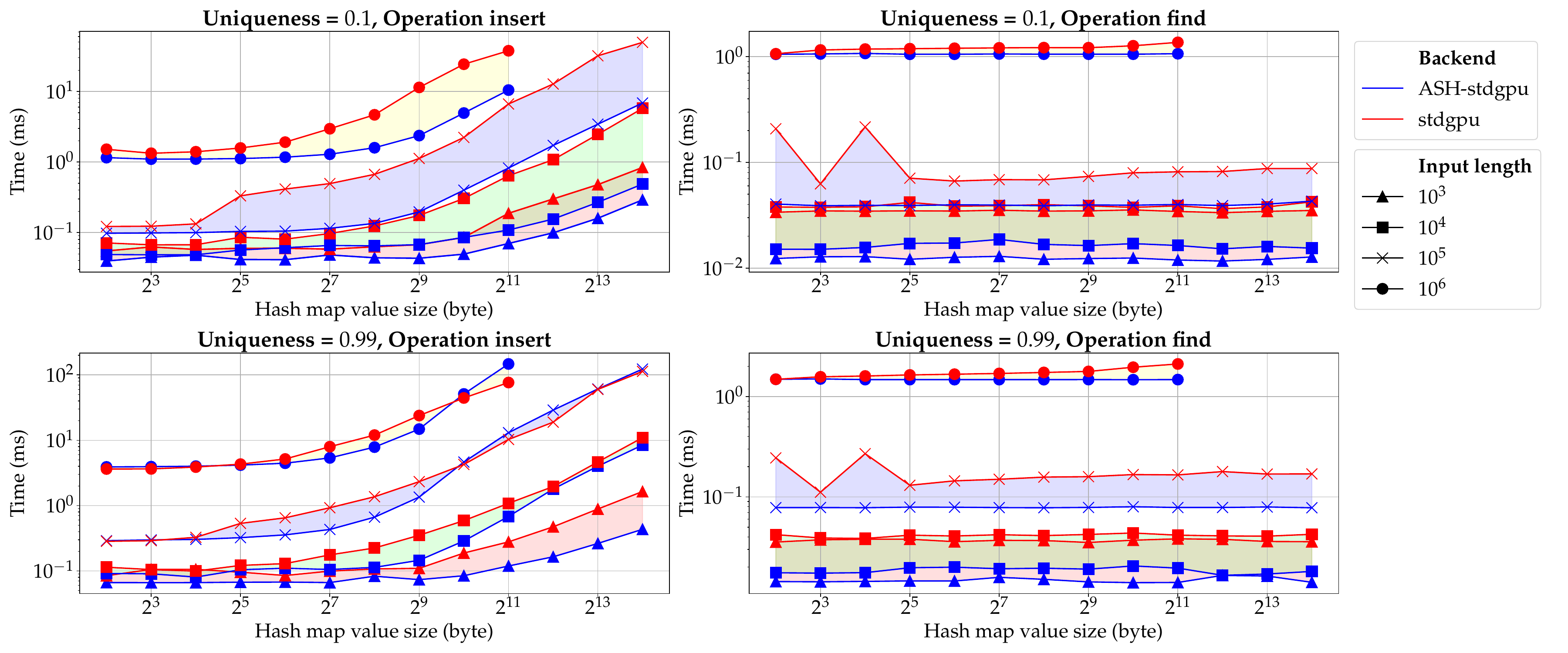}
    \caption{Hash map performance comparison between ASH-stdgpu and the vanilla stdgpu with 3D integer keys. Each curve shows the average operation time (\emph{y-axis}) with varying hash map value sizes in bytes (\emph{x-axis}), given a controlled backend, input length, and input key uniqueness ratio. \emph{Lower is better}.
    Further factors are denoted by the legends on the right. ASH-stdgpu runs consistently faster than the vanilla stdgpu.}
    \label{fig:exp-stdgpu}
\end{figure*}

We start with synthetic experiments to show that ASH, with its optimized memory layout, increases performance while improving usability. All experiments in this section are conducted on a laptop with an Intel i7-6700HQ CPU and an Nvidia GTX 1070 GPU. In all experiments, we assume the hash map capacity is equivalent to the number of input keys (regardless of duplicates). Each reported time is an average of 10 trials.

\subsection{Spatial Hashing with Generic Backend}\label{subsec:exp-stdgpu}
The first experiment is the performance comparison between vanilla stdgpu and ASH with stdgpu backend (ASH-stdgpu). For fairness, we extend the examples of stdgpu such that an array of iterators and masks are returned for in-place manipulations. The number of buckets and the load factor are determined internally by stdgpu.

\begin{setup}
    We test randomly generated 3D spatial coordinates mapped to float value blocks of varying sizes. The key $\kK$, value $\vV$, capacity $c$, and uniqueness $\rho$ are chosen as follows:
    \begin{equation*}
        \begin{split}
            \kK & = \{\normaltt{Tensor((3), Int32)}\}, \\
            \vV & = \{\normaltt{Tensor((}\mathtt{2^j}\normaltt{), Float32)} \mid j=0,1,\dotsc,12 \}, \\
            c   & = \{10^j \mid j=3, 4, 5, 6\}, \\
            \rho   & = \{0.1, 0.99\},
        \end{split}
    \end{equation*}
    where $\rho$ indicates the ratio of the unique number of keys to the total number of keys being inserted or queried.
    \label{setup:int3-stdgpu}
\end{setup}
Fig.~\ref{fig:exp-stdgpu} illustrates the comparison between vanilla stdgpu and ASH-stdgpu. For \emph{insert} operation, ASH-stdgpu is significantly faster than stdgpu when $\rho=0.1$ is low, and the performance gain increases when the value byte size increases. This is mainly due to the SoA memory layout and the lazy insertion mechanism, where a lightweight integer $i \in \iI_c$ is inserted in an attempt instead of the actual value $v \in \vV$. At a high input uniqueness $\rho=0.99$, ASH-stdgpu maintains the performance advantage with low and medium value sizes, and its performance is comparable to stdgpu with a large value size. This indicates that our dispatch pattern in copying values helps in a high throughput scenario. For \emph{find} operation, ASH-stdgpu is consistently faster than vanilla stdgpu, under both high and low key uniqueness settings.

In addition to to \emph{insert} and \emph{find}, we introduce a new \emph{activate} operation. It ``activates'' the input keys by inserting them into the hash map and obtaining the associated buffer indices. This is especially useful when we can pre-determine and apply the element-wise initialization. Examples include the TSDF voxel blocks (zeros) and multi-layer perceptrons (random initializations). The \emph{activate} operation is absent in most existing hash maps and is only available as hard-coded functions~\cite{niessner2013real,prisacariu2017infinitam, dong2019gpu}.

\begin{figure*}[ht]
    \centering
    \includegraphics[width=\linewidth]{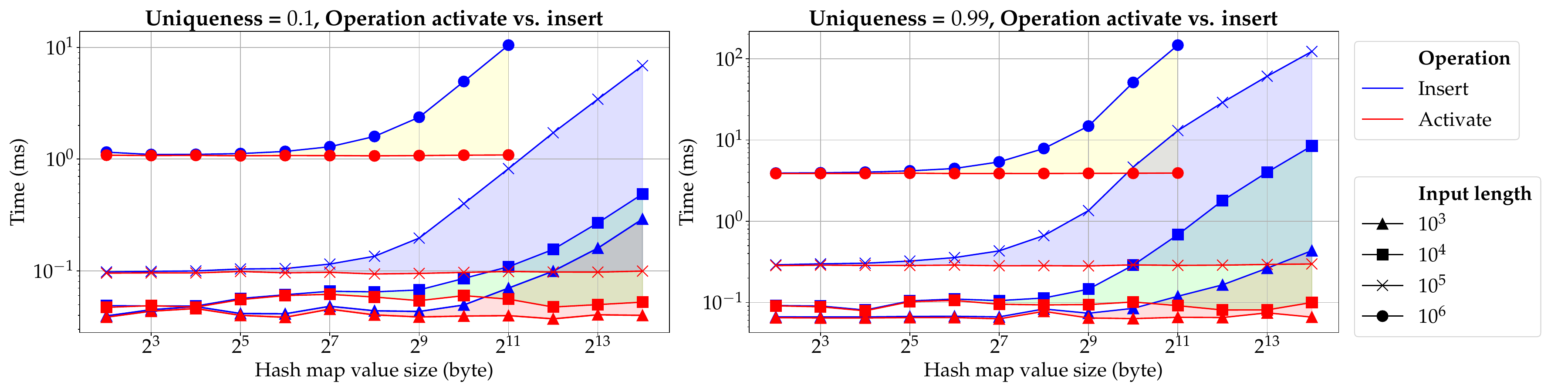}
    \caption{Study of the \emph{activate} operation introduced in ASH against \emph{insert} with 3D integer keys on the ASH-stdgpu backend. Each curve shows the average operation time (\emph{y-axis}) with varying hash map value sizes in bytes (\emph{x-axis}), given an input length and input key uniqueness ratio. \emph{Lower is better}.
    \emph{Activate} keeps a stable runtime in the tasks that do not require explicit value insertion, while \emph{insert} time increases corresponding to the hash map value size. }
    \label{fig:exp-ablation-activate}
\end{figure*}

With the \emph{activate} operation, we conduct ablation studies to compare the insertion time of merely the keys versus the insert time of both the keys and values. Fig.~\ref{fig:exp-ablation-activate} compares the runtime between \emph{insert} and \emph{activate} in ASH-stdgpu. The key, value, capacity, and uniqueness choices are the same as in Setup~\ref{setup:int3-stdgpu}. We observe that while the insertion time increases as the value size increases, the activation time remains stable.

\subsection{Integer Hashing with Non-Generic Backend}

Next, we compare ASH based on the SlabHash backend (ASH-slab) with the vanilla SlabHash. Since SlabHash only supports integers as keys and values, we limit our ASH-slab backend to the same integer types here. The number of buckets is $2\times$ capacity (load factor is approx. $0.5$), since it is empirically the best factor when ASH-slab is applied to non-generic and generic tasks. Since vanilla SlabHash only supports data I/O from the host, we include the data transfer time between host and device when measuring the performance of ASH-slab.

\begin{setup}
    We test random scalar integer values mapped to scalar float values. The key $\kK$, value $\vV$, capacity $c$, and uniqueness $\rho$ are chosen as follows:
    \begin{equation*}
        \begin{split}
            \kK & = \{\normaltt{Tensor((1), Int32)}\}, \\
            \vV & = \{\normaltt{Tensor((1), Float32)}\}, \\
            c   & = \{10^j \mid j=3, 4, 5, 6\}, \\
            \rho   & = \{0.1, 0.2, \dotsc, 0.9, 0.99\},
        \end{split}
    \end{equation*}
\end{setup}

\begin{figure*}[ht]
    \centering
    \includegraphics[width=\linewidth]{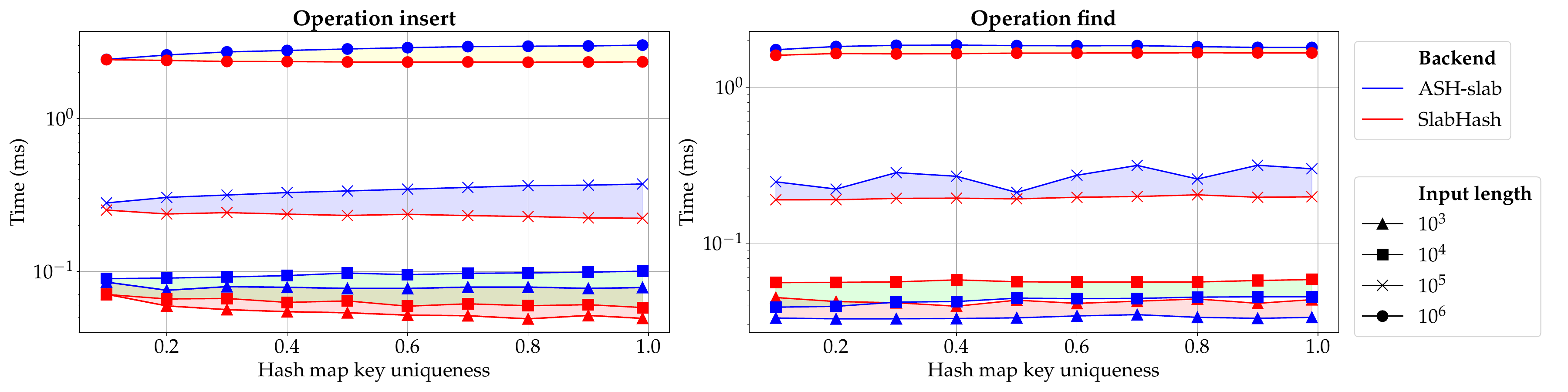}
    \caption{Hash map performance comparison between ASH-slab and the vanilla SlabHash with 1D integer keys and values. Each curve shows the average operation time (\emph{y-axis}) with varying input key uniqueness ratio (\emph{x-axis}), given an input length. \emph{Lower is better}.
    ASH-slab retains a comparable performance for integers while supporting generalization to arbitrary dimensional keys and values of various data types.}
    \label{fig:exp-slab}
\end{figure*}

As shown in Fig.~\ref{fig:exp-slab}, although ASH-slab does not make use of the non-blocking warp-oriented operations in SlabHash in order to enable support for generic key and value types, our \emph{insert} is still comparable to vanilla SlabHash which is only optimized for integers. The drop in performance of ASH-slab when $\rho$ increases is an expected indication that the overhead of multi-pass insertion increases correspondingly.

It is worth mentioning that with an improved global memory manager, the construction of an ASH-slab hash map takes less than 1ms under all circumstances, while the vanilla SlabHash constantly takes 30ms for the redundant slab memory manager. In practice, where the hash map is constructed and used once (\eg~ voxelization), ASH-slab is a more practical solution.

\subsection{Ablation Between Backends}
We now conduct an ablation study with different backends in ASH, namely ASH-stdgpu and ASH-slab, with arbitrary input key-value types beyond integers. The experimental setup follows Section~\ref{subsec:exp-stdgpu}.
\begin{figure*}[ht]
    \centering
    \includegraphics[width=\linewidth]{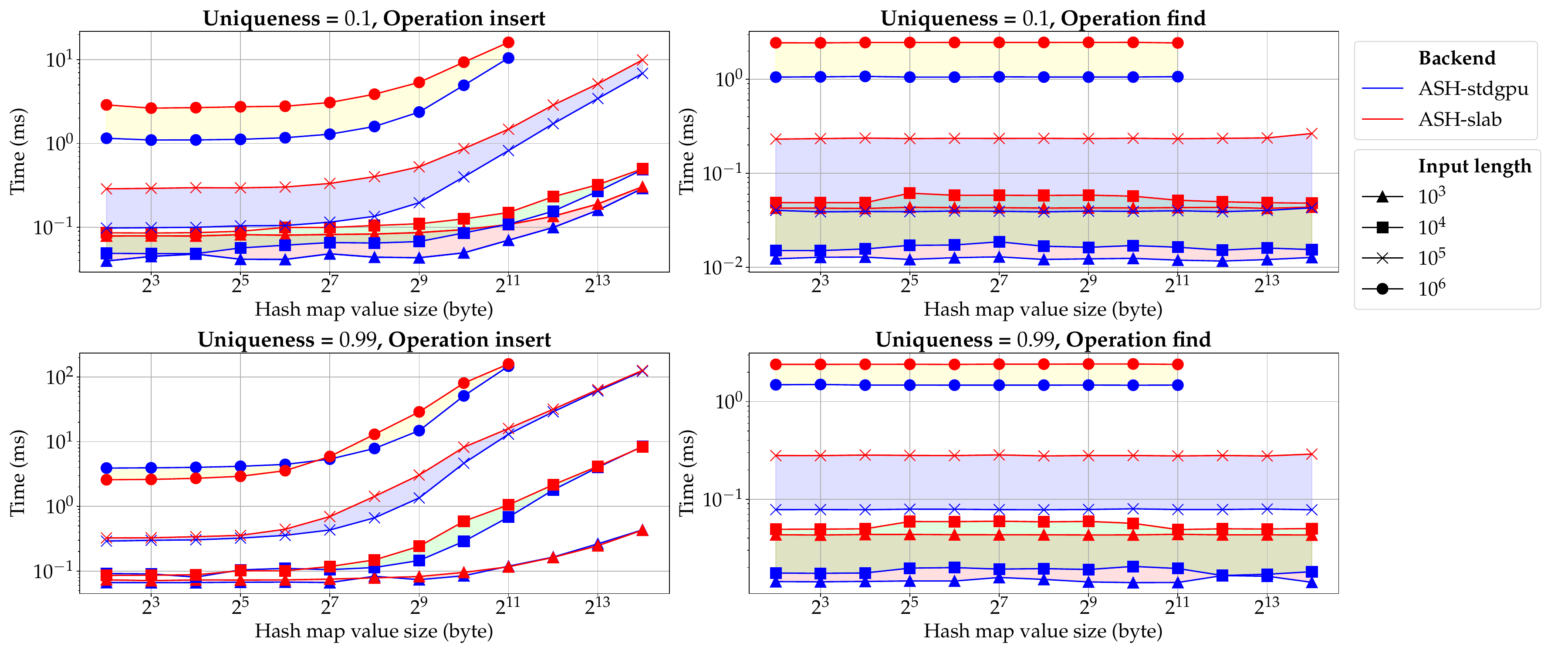}
    \caption{Ablation study of the hash map performance with 3D integer keys over different backends. Each curve shows the average operation time (\emph{y-axis}) with varying hash map value sizes in bytes (\emph{x-axis}), given a controlled backend, input length, and input key uniqueness ratio. \emph{Lower is better}.
    ASH-stdgpu outperforms ASH-slab in most circumstances with the 3D integer keys and varying length values, which are common in real-world applications.}
    \label{fig:exp-ablation-backend}
\end{figure*}

In Fig.~\ref{fig:exp-ablation-backend}, we can see that ASH-stdgpu outperforms ASH-slab in most circumstances with the 3D coordinate keys and varying length values that are common in real-world applications. While warp-oriented operations heavily used in SlabHash enjoy the benefits of intrinsic acceleration, they sacrifice the granularity of operations. Threads can only move on to the next task once all the operations in a warp (of 32) are finished. As a result, early termination when an insertion failure occurs are less likely in a warp-oriented hash map. If the data layout is not well-distributed for the intrinsic operations (\eg, low-uniqueness input, keys with long word width), the performance drop could be significant.

This observation is more apparent in insertion under varying input densities. With a relatively small value size and a high uniqueness, ASH-slab performs better. When the uniqueness is low, however, each thread in ASH-slab still has to finish a similar workload before termination, while ASH-stdgpu can reject many failure insertions early and move on to the following workloads. As of now, ASH-slab is suitable for the voxel downsampling application, while ASH-stdgpu is better for other tasks. Therefore, we set stdgpu as the default backend for ASH in the remaining sections.

\subsection{Code Complexity}
We now study the usage at the user end. First of all, the ASH framework, regardless of the backend used, is already compiled as a library. A C++ developer can easily include the header and build the example directly with a CPU compiler and link to the precompiled library with a light tensor engine. An equivalent Python interface is provided via pybind~\cite{pybind11} as shown in Fig.~\ref{fig:hash-map-interface-ash}.

In comparison, to use SlabHash's interface with an input array from host memory, a CUDA compiler is required, along with manual bucket-linked list chain configurations. For further performance improvement, detailed memory management has to be done manually via {\ttfamily cudaMalloc, cudaMemcpy} and {\ttfamily cudaFree}. stdgpu provides a built-in memory manager but requires writing device and host functions.
In a query operation, the found values are returned \emph{by-copy} for SlabHash, so in-place modification requires further modification of the library. stdgpu exposes iterators in an AoS fashion, therefore the device code needs to be implemented to reinterpret an array of iterators and masks for further operations.

The compilation complexity and the interface LoC required for the same functionality in C++ are listed in Table~\ref{tab:exp-comparison-loc}.
\begin{table}
    \caption{Comparison of the complexity of coding (top) and LoC (bottom) of each operation among the implementations. Unlike stdgpu and SlabHash, ASH does not require that users write device code or use a CUDA compiler. It requires few LoC for construction, query, and insertion.}
    \centering
    \ra{1.05}
    \small
    \resizebox{0.8\linewidth}{!}{
    \begin{tabular}{@{}lccc@{}}
      \toprule
      & {stdgpu} & {SlabHash} & {ASH}\\
      \midrule
      Device code free? & \NoX & \YesV & \YesV\\
      CUDA compiler free? & \NoX & \NoX & \YesV\\
      \midrule
      Construct & 3 & 9 & 3\\
      Find & 22 & 2 & 2\\
      Insert & 27 & 1 & 2\\
      \bottomrule
    \end{tabular}
    }
    \label{tab:exp-comparison-loc}
\end{table}

%% file: tex/application.tex
We now demonstrate a number of applications and ready-to-use systems in 3D perception to demonstrate the power of ASH with fewer LoC and better performance.
The presented applications include:
\begin{enumerate}
    \item Point cloud voxelization;
    \item Retargetable volumetric reconstruction;
    \item Non-rigid registration and deformation;
    \item Joint geometry and appearance refinement.
\end{enumerate}
The first two experiments are conducted on an Intel i7-6700HQ CPU and an Nvidia GeForce GTX 1070 GPU for indoor scenes. Outdoor scene experiments are run on an Intel i7-11700 CPU and an Nvidia RTX 3060 GPU. The rest are done on an Intel i7-7700 CPU and an Nvidia GeForce GTX 1080Ti GPU.

\input{tex/apps/voxelization.tex}
\input{tex/apps/voxelhashing.tex}
\input{tex/apps/slac.tex}
\input{tex/apps/sfs.tex}

%% file: tex/apps/voxelization.tex
\begin{figure}
  \centering
  \begin{tabular}{@{}c@{\hspace{1mm}}c@{}}
    \includegraphics[width=0.48\linewidth]{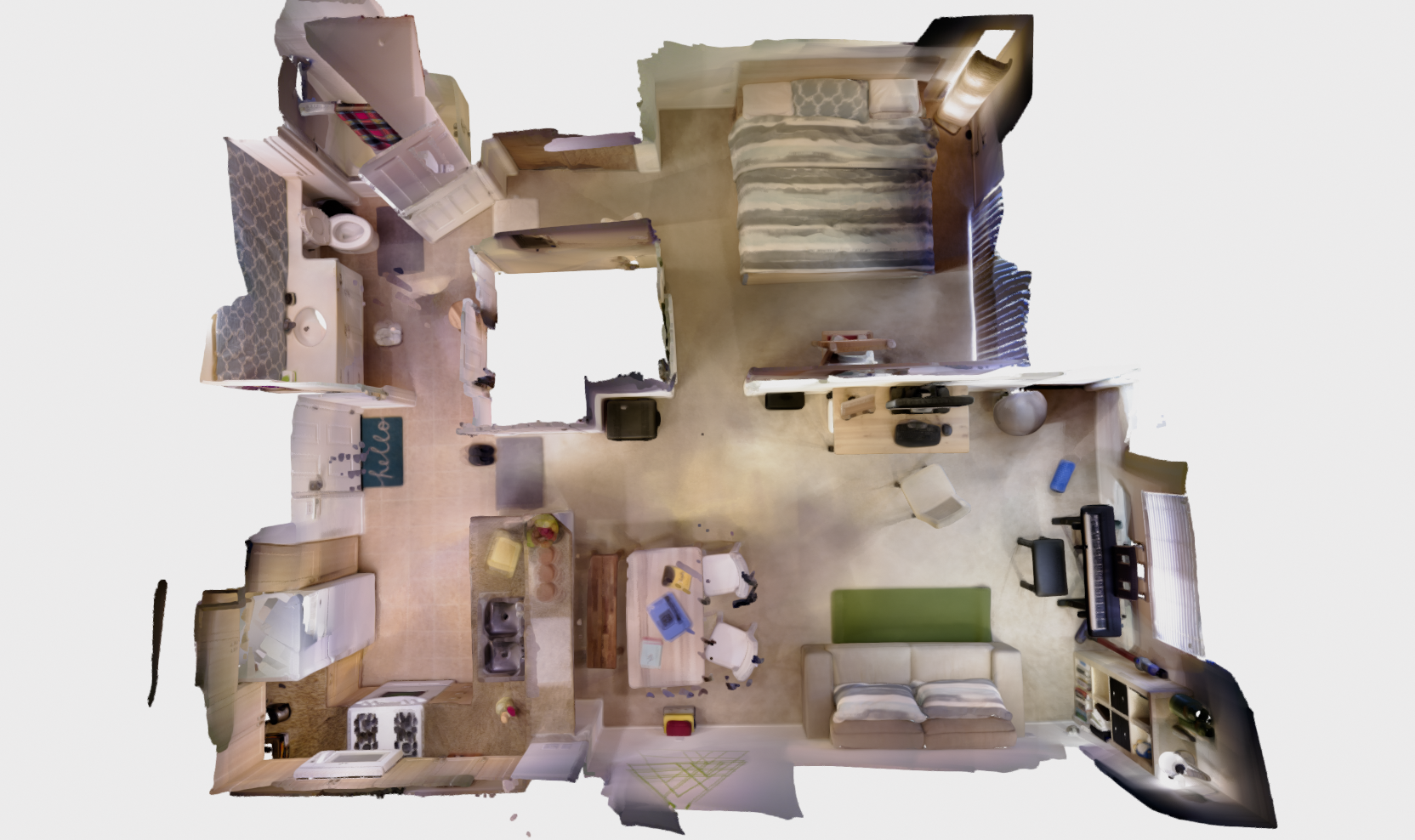} &
    \includegraphics[width=0.48\linewidth]{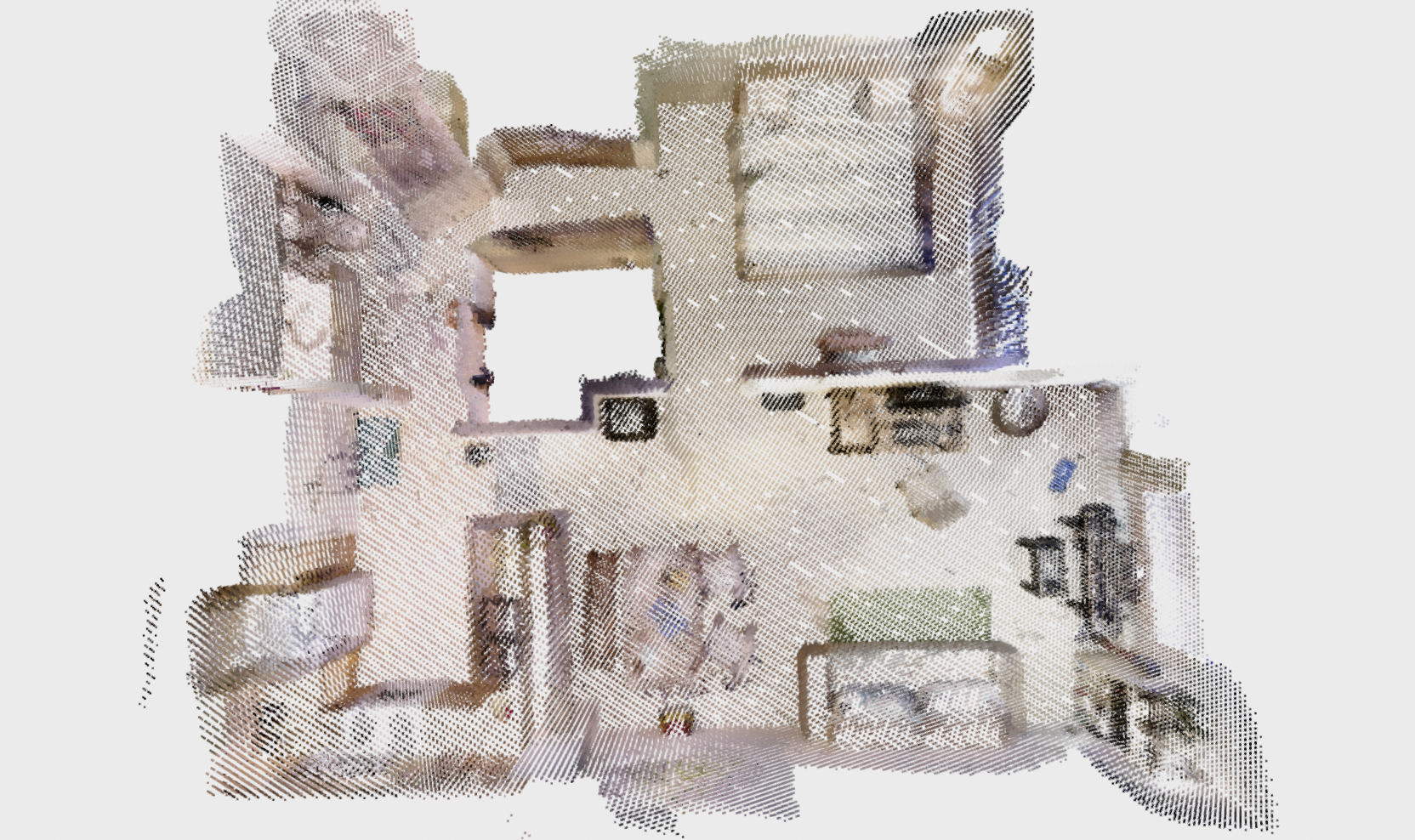} \\
    \includegraphics[width=0.48\linewidth]{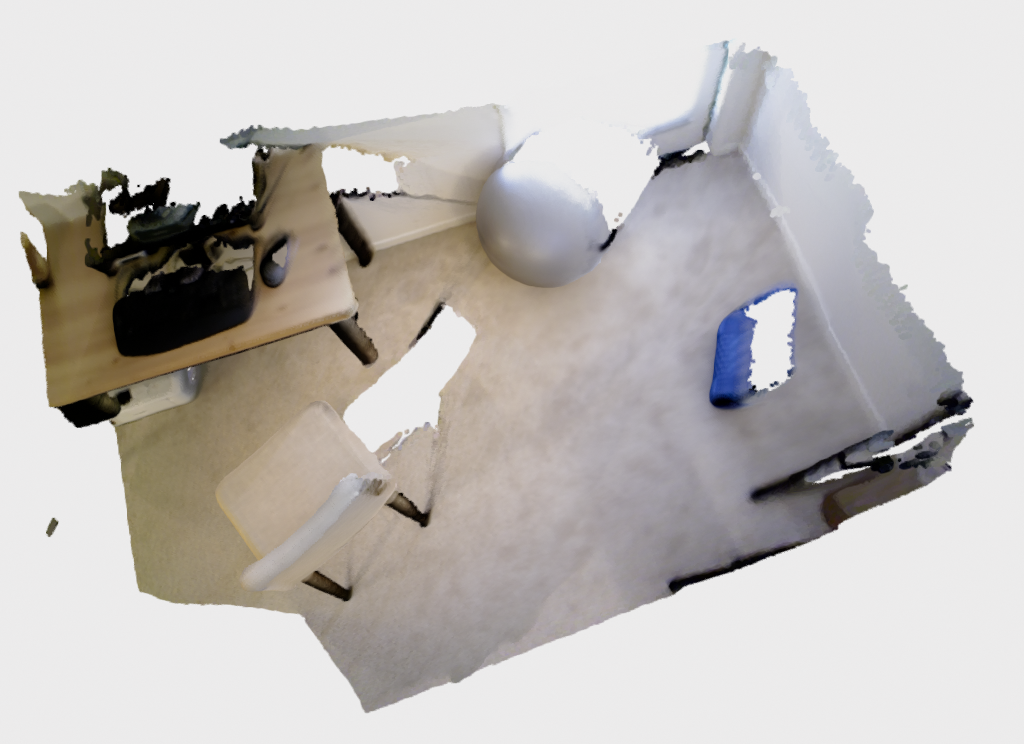} &
    \includegraphics[width=0.48\linewidth]{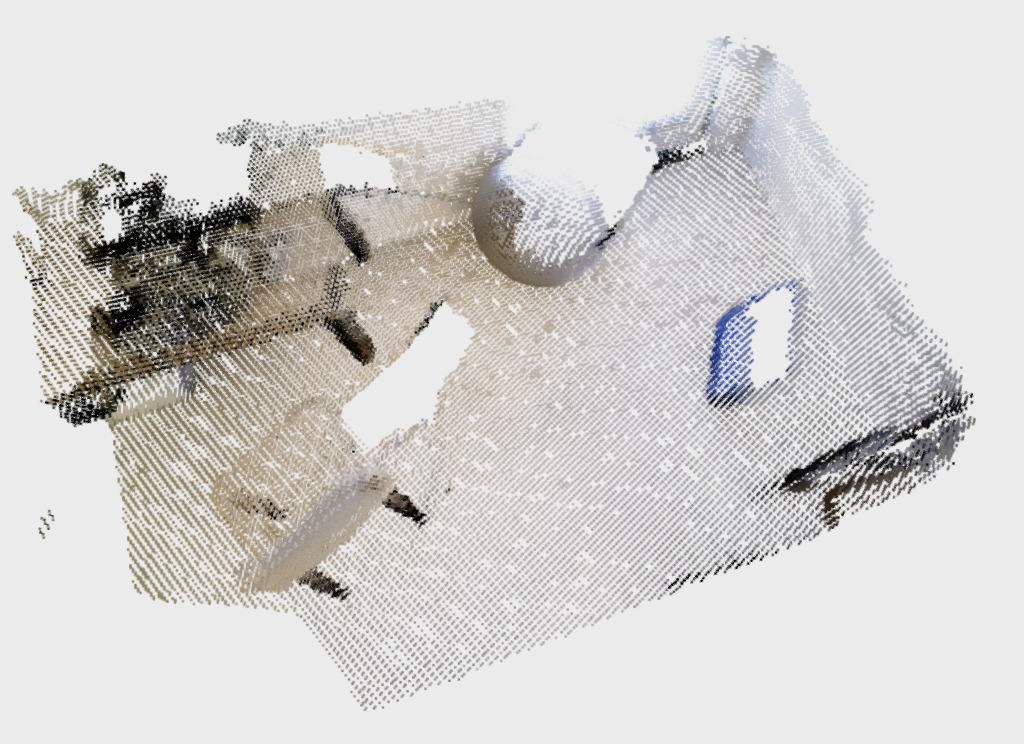}
  \end{tabular}
  \caption{Visualization of point cloud voxelization. Top: scene-level large-scale inputs. Bottom: fragment-level small-scale inputs. Left: original point clouds. Right: voxelized point clouds.}
  \label{fig:app-voxelization-pcd}
\end{figure}

\subsection{Point Cloud Voxelization}
\begin{setup}
    In voxelization, a hash map maps a point cloud's discretized coordinates to its natural array indices, and the hash map capacity is the point cloud size, typically ranging from $10^5$ to $10^7$
    \begin{equation*}
        \begin{split}
            \kK & = \{\normaltt{Tensor((3), Int32)}\}, \\
            \vV & = \{\normaltt{Tensor((1), Int32)}\}.
        \end{split}
    \end{equation*}
\end{setup}

Voxelization is a core operation for discretizing coordinates. It is essential for sparse convolution~\cite{choy20194d, choy2020deep} at the quantization preprocessing stage, and is often used to generate point cloud ``pyramids''~\cite{zhou2018open3d} in coarse-to-fine 3D registration for improved speed and robustness.

Voxelization is a natural task for parallel hashing, as the essence of the operation is to discard duplicates at grid points. To achieve this, we first discretize the input by converting the coordinates described by the continuous meter metric to the voxel units. Then a simple hash set insertion eliminates the duplicates and corresponds them to the remaining unique coordinates.
The returned indices can be reused for tracing other properties such as colors and point normals associated with the input. The python code for voxelization can be found in Listing~\ref{alg:voxelization}.
\vspace{-2mm}
\begin{lstlisting}[language=Python, caption={Voxelization}, label={alg:voxelization}]
  import open3d.core as o3c
  
  def voxelize(pcd, voxel_size):
      xyz = pcd.point.positions
      N = len(xyz) 
      hashset = o3c.HashSet(N, o3c.int32, (3,))
      xyz_int = (xyz / voxel_size).floor().to(o3c.int32)
      _, mask = hashset.insert(xyz_int)
      # Return points with indices
      return xyz[mask], o3c.arange(N)[mask]
\end{lstlisting}
\vspace{-2mm}

We compare voxelization implemented in ASH with two popular implementations, MinkowskiEngine~\cite{choy20194d} on CUDA and Open3D~\cite{zhou2018open3d} on CPU. Our experiments are conducted on a large scene input with $8 \times 10^6$ points which is typical for scene perception, and a small fragment of the scene with $5 \times 10^5$ points which is typical for an RGB-D input frame, as shown in Fig.~\ref{fig:app-voxelization-pcd}.

\begin{figure}[ht]
    \includegraphics[width=\linewidth]{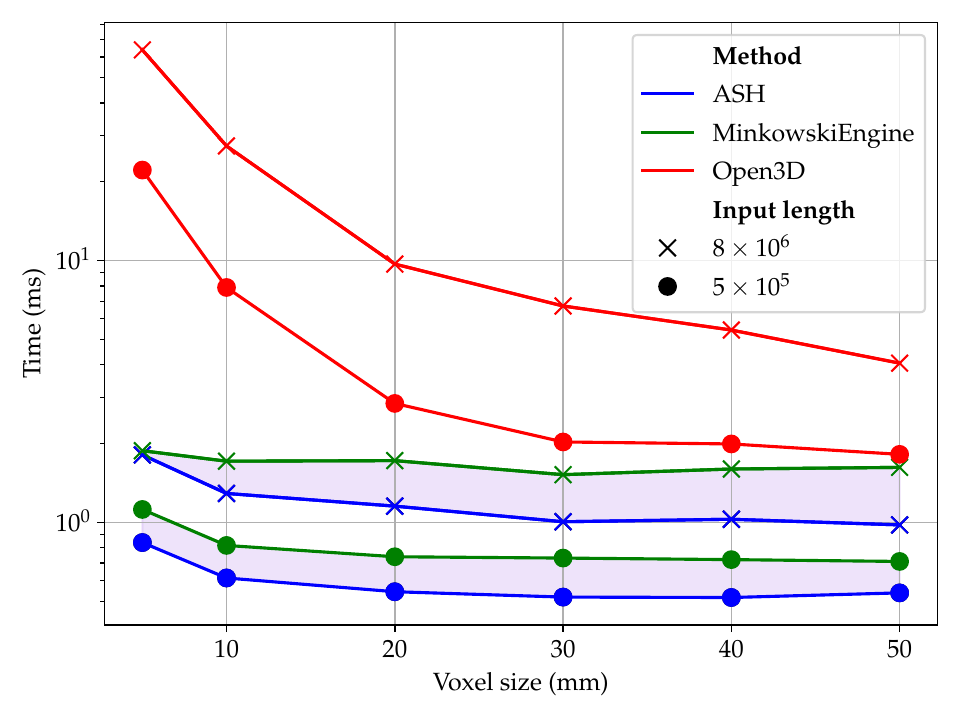}
    \caption{Performance comparison of voxelization. Each curve shows the run time (\emph{y-axis}) over the varying voxel size (\emph{x-axis}). \emph{Lower is better}.
    ASH is consistently faster than Open3D's default voxelizer (CPU) and MinkowskiEngine (CUDA).}
    \label{fig:app-voxelization-performance}
\end{figure}

To evaluate the performance, we vary the parameter \emph{voxel size} from 5mm to 5cm, which is typical in the spectrum of voxelization applications, from dense reconstruction to feature extraction. In Fig.~\ref{fig:app-voxelization-performance} we can see that our implementation outperforms baselines consistently for inputs at both scales. Meanwhile, in measuring the LoC written in C++ (the Python wrappers are one-liners) required for the functionality given the hash map interface, we observe that ASH requires only 28 LoC, while MinkowskiEngine and Open3D on CPU take 71 and 72 LoC respectively.

%% file: tex/apps/voxelhashing.tex
\subsection{Retargetable Volumetric Reconstruction}

\subsubsection{Truncated Signed Distance Function}
Scene representation with truncated signed distance function (TSDF) from a sequence of 3D input has been introduced~\cite{curless1996volumetric} and adapted to RGB-D~\cite{newcombe2011kinectfusion}.
It takes a sequence of depth images $\{D^j\}$ with their poses $\{\TT^j \in \mathbf{SE}(3)\}$ as input, and seeks to optimize the signed distance, an implicit function value $d$ per point at $\xx \in \mathbb{R}^3$. The signed distance measured for frame $j$ is given by\footnote{Details including depth masking and projective pinhole camera model are omitted for clarity and could be found in KinectFusion~\cite{newcombe2011kinectfusion}.}
\begin{align}
    [u, v, r]^\top &= \Pi({\TT^j}^{-1} \xx),\\
    d^j &= D^j(u, v) - r,
\end{align}
where $\Pi$ projects the 3D point to 2D with a range reading after a rigid transformation. To reject outliers, a truncate function $\Psi_\mu(d) = \mathrm{clamp}(d, -\mu, \mu)$ is applied to $d^j$.
There are multiple variations of $\Psi$ and the definition of signed distance $d^j$~\cite{bylow2013direct}.
For this paper, we follow the convention in KinectFusion~\cite{newcombe2011kinectfusion}.

With a sequence of inputs, per-point signed distance can be estimated in least squares with a closed-form solution
\begin{align}
    d &= \argmin_{t} \sum_{j} w^j \lVert t - d^j \lVert^2,
    d = \frac{ \sum_{j} w^j d^j } { \sum_j w^j } \label{equ:integration-batch},
\end{align}
where $w^j$ is the selected weight depending on view angles and distances~\cite{bylow2013direct}. In other words, with a sequence of depth inputs and their poses, we can measure TSDF at any point in 3D within the camera frustums. We can also rewrite Eq.~\ref{equ:integration-batch} incrementally:
\begin{align}
    d &:= \frac{ w \cdot d + w^j \cdot d^j } { w + w^j },     w := w + w^j,
    \label{equ:integration-inc}
\end{align}
where $w$ is the accumulated weight paired with $d$.

Equipped with a projection model $\Pi$ that converts a point to signed distance, TSDF reconstruction can be generalized to imaging LiDARs~\cite{dong2021lidar} for larger scale scenes.

\subsubsection{Spatially Hashed TSDF Blocks}
\begin{setup}\label{setup:voxelblock}
    In a scene represented by a volumetric TSDF grid, a hash map maps the coarse voxel blocks' coordinates to the TSDF data structure of the voxel block, and the hash map capacity is typically $10^3$ to $10^5$ for small to large-scale indoor scenes:
    \begin{equation*}
        \begin{split}
            \kK & = \{\normaltt{Tensor((3), Int32)}\}, \\
            \vV & = \{\normaltt{Tensor((}\ell^3\normaltt{), Float32)}, \normaltt{Tensor((}\ell^3\normaltt{), Float32)}\},
        \end{split}
    \end{equation*}
    where $\ell$ is the voxel block resolution, which is set to $8$ or $16$.
\end{setup}
While recent neural representations utilize multi-layer perceptrons to approximate the TSDF in continuous space~\cite{chabra2020deep}, classical approaches use discretized voxels. Such representations have a long history and are ready for real-world, real-time applications. They can also provide data for training neural representations.

The state-of-the-art volumetric discretization for TSDF reconstruction is spatial hashing, where points are allocated around surfaces on-demand at a voxel resolution of around $5mm$.
While it is possible to hash high-resolution voxels directly, the access pattern could be less cache-friendly, as the neighbor voxels are scattered in the hash map. A hierarchical structure is a better layout, where small dense voxel grids (\eg~in the shape of $8^3$ or $16^3$) are the minimal unit in a hash map; detailed access can be redirected to simple 3D indexing. In other words, a voxel can be indexed by a coupled hash map lookup and a direct local addressing
\begin{align}
    \xx_{\text{block}} &= \lfloor \xx / (s \ell) \rfloor,\label{equ:block-coordinate}\\
    \xx_{\text{voxel}} &= \lfloor (\xx - \xx_{\text{block}} \cdot s \ell) / s \rfloor\label{equ:voxel-coordinate},
\end{align}
where $s$ is the voxel size and $\ell$ is the voxel block resolution as described in Setup~\ref{setup:voxelblock}.

While previous implementations~\cite{niessner2013real,prisacariu2017infinitam,dong2019gpu} have achieved remarkable performance, modularized designs are missing. Geometry processing and hash map operations were coupled due to the absence of a well-designed parallel GPU hash map. One deficiency of this design is \emph{unsafe} parallel insertion, where the capacity of a hash map can be exceeded. Another is ad hoc recurring low-level linked list access in geometry processing kernels that cause high code redundancy. Our implementation demonstrates the first modularized pipeline where safe hash map operations are used without any ad hoc modifications.

\subsubsection{Voxel Block Allocation and TSDF Integration}
\begin{setup}\label{setup:local}
    For an input depth image, a hash map maps the unprojected point coordinates  to the active indices as described in Setup~\ref{setup:voxelblock}, and the  capacity of a hash map is typically $10^5$ to $10^6$, with $10^2$ to $10^3$ valid entries:
    \begin{equation*}
        \begin{split}
            \kK & = \{\normaltt{Tensor((3), Int32)}\}, \\
            \vV & = \{\normaltt{Tensor((1), Int32)}\}.
        \end{split}
    \end{equation*}
\end{setup}

In the modularized design, we first introduce a double hash map structure for voxel block allocation and TSDF estimation. Voxel block allocation identifies points from $\{D^j\}$ as surfaces and computes coordinates with Eq.~\ref{equ:block-coordinate}. Intuitively, they can be directly inserted to the \emph{global} hash map described in Setup~\ref{setup:voxelblock}. This is achievable in an ad hoc implementation where the core of the hash map is modified at the device code level, and \emph{unsafe} insertion is allowed~\cite{niessner2013real,prisacariu2017infinitam}.
However, in a modularized and \emph{safe} setup, this could lead to problems. A VGA resolution depth input contains $640 \times 320 \approx 3 \times 10^5$ points and easily exceeds the empirical \emph{global} hash map capacity. As we have mentioned, rehashing will be triggered under such circumstances, which is both time and memory consuming, especially for a hash map with memory-demanding voxel blocks as values.

To address this issue without changing the low-level implementation and sacrificing safety, we introduce a second hash map, the \emph{local} hash map from Setup~\ref{setup:local}. This hash map is similar to the one used in \emph{voxelization}: it maps discretized 3D coordinates unprojected from depths to integer indices. With this setup, a larger input capacity is acceptable, as the \emph{local} hash map is lightweight and can be cleared or constructed from scratch per iteration.

\begin{figure}[ht]
    \centering
    \includegraphics[width=0.9\linewidth]{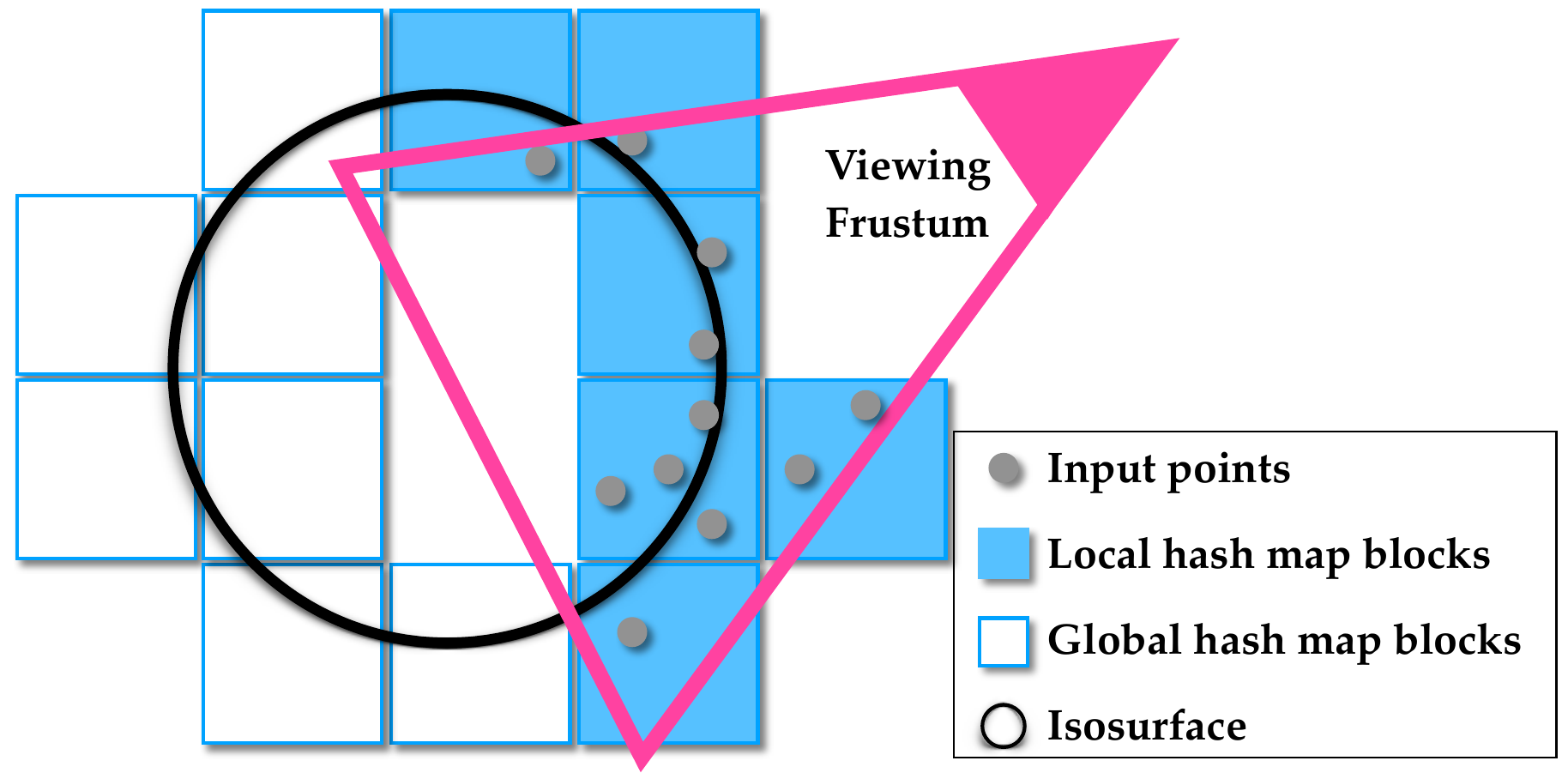}
    \caption{Illustration of local and global hash maps iteratively used in real-time reconstruction.
    The local hash map activates voxel blocks enclosing points observed in the viewing frustum. The global hashmap accumulates such activated blocks and maintains all the blocks around the isosurface.}
    \label{fig:app-integration-illustration}
\end{figure}

There are two main benefits to using a \emph{local} hash map: it converts the input from the $10^5$ raw point scale to the $10^3$ voxel block scale, which is safe for the \emph{global} hash map without rehashing; as a byproduct, it keeps track of the active voxel blocks for the current frame $j$, which can be directly used in the following TSDF integration and ray casting. The local and global hash maps can be connected through indices, where a query of coordinate in the local map is redirected to the global map in-place. Fig.~\ref{fig:app-integration-illustration} shows the roles of the two hash maps. Listing~\ref{alg:double-hash-map} details the construction and interaction between the two hash maps.

\vspace{-2mm}
\begin{lstlisting}[language=Python, caption={Double hash map allocation}, label={alg:double-hash-map}]
  import open3d.core as o3c
  
  # Map block coords to actual storage
  global_hashmap = o3c.HashMap(
      global_capacity,
      key_dtype=o3c.int32,
      key_shape=(3,),
      # Float 1-channel TSDF and 3-channel color
      value_dtypes=(o3c.float32, o3c.float32),
      values_shapes=((8, 8, 8, 1), (8, 8, 8, 3)))
  # Map block coords to global hashmap indices
  local_hashmap = o3c.HashMap(
      local_capacity,
      key_dtype=o3c.int32,
      key_shape=(3,)
      # Index in global hash map
      value_dtypes=(o3c.int32),
      value_shapes=((1,)))
  # Discretize and insert to local map
  xyz_int = (xyz / block_size).floor().to(o3c.int32)
  i_local, mask = local_hashmap.activate(xyz_int)
  # Remove duplicates
  xyz_int = xyz_int[mask]
  i_local = i_local[mask]
  # Activate and query in the global map
  global_hashmap.activate(xyz_int)
  i_global, mask = global_hashmap.find(xyz_int)
  # Associate local and global maps via indices
  local_v = local_hashmap.value()
  local_v[i_local] = i_global
\end{lstlisting}
\vspace{-2mm}

By accessing the global hash map's TSDF and colors through returned indices, TSDF integration can then be implemented following Eq.~\ref{equ:integration-inc} in a pure geometry function, either in a low-level GPU kernel or a high-level vectorized Python script. Spatial hash map is detached from the core geometric computation, providing more flexibility in performance optimization.

\subsubsection{Surface Extraction}
A volumetric scene reconstruction is not usable for most software and solutions until the results are exported to point clouds or triangle meshes.
Hence we implement a variation of Marching Cubes~\cite{lorensen1987marching} that extracts vertices with triangle faces at zero crossings in a volume. In a spatially hashed implementation, boundary voxels in voxel blocks are harder to process since queries of neighbor voxel blocks are frequent, and shared vertices among triangles are hard to track.
One common approach is to simply visit vertices at isosurfaces and disregard duplicates, but this usually results in a heavily redundant mesh~\cite{klingensmith2015chisel,prisacariu2017infinitam}, or time-consuming post-processing to merge vertices~\cite{niessner2013real}.
Another method is to introduce an assistant volumetric data structure to atomically record the vertex-voxel association, but the implementations are over-complex and require frequent low-level hash map queries coupled with surface extraction~\cite{dong2018efficient, dong2019gpu}.

Now that we have a unified hash map interface, we simplify the voxel block neighbor search routine~\cite{dong2019gpu} and set up a \emph{1-radius neighbor} lookup table in advance, as described in Listing~\ref{alg:one-radius-nn}. Surface extraction is then detached from hash map access and can be optimized separately.
As a low-hanging fruit, point cloud extraction is implemented with the same routine by ignoring the triangle generation step. In fact, surface extraction of a median-scale scene shown in Fig.~\ref{fig:app-realtime-qualitative} takes less than 100ms, making interactive surface updates possible in a real-time reconstruction system.

\vspace{-2mm}
\begin{lstlisting}[language=Python, caption={Radius nearest neighbor search in 3D}, label={alg:one-radius-nn}]
  import open3d.core as o3c
  
  def radius_nns(xyz, r):
      N = len(xyz)
      hashset = o3c.HashSet(N, o3c.int32, (3,))
      hashset.insert(xyz)
      # Get offset tensors
      # ([-r, -r, -r], ..., [r, r, r])
      offsets = enumerate_radius_offsets(r)
      # Collect neighbors
      xyz_query = xyz.clone()
      for offset in offsets:
          xyz_query.append(xyz + offset, 0)
      # Query
      indices, masks = hashset.find(xyz_query)
      # Reshape to get neighbor indices for each point
      indices = indices.view(N, -1)
      masks = indices.view(N, -1)
\end{lstlisting}
\vspace{-2mm}

\subsubsection{Ray Casting}
Another way to interpret a volume is through ray casting or ray marching.
Given camera intrinsics and extrinsics, ray casting renders depth and color images by marching rays in the spatially hashed volumes, querying color and TSDF values, and finding zero-crossing interfaces. It allows rendering at known viewpoints, synthesizing novel views, and estimating camera poses.

\begin{figure}[ht]
  \centering
  \begin{tabular}{@{}c@{\hspace{1mm}}c@{\hspace{1mm}}c@{}}
    \includegraphics[height = 2.2cm]{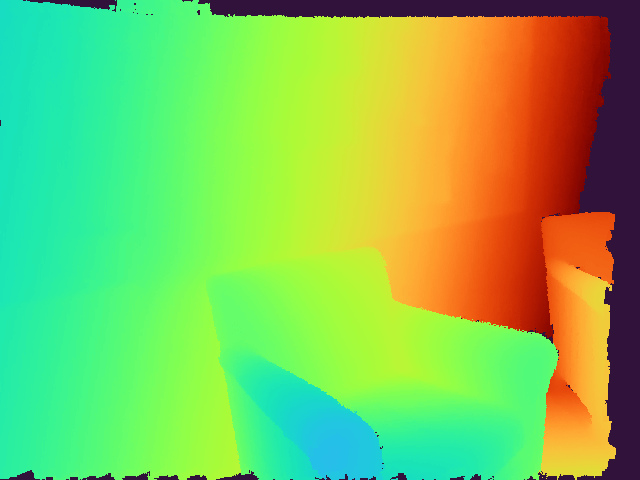} & 
    \includegraphics[height = 2.2cm]{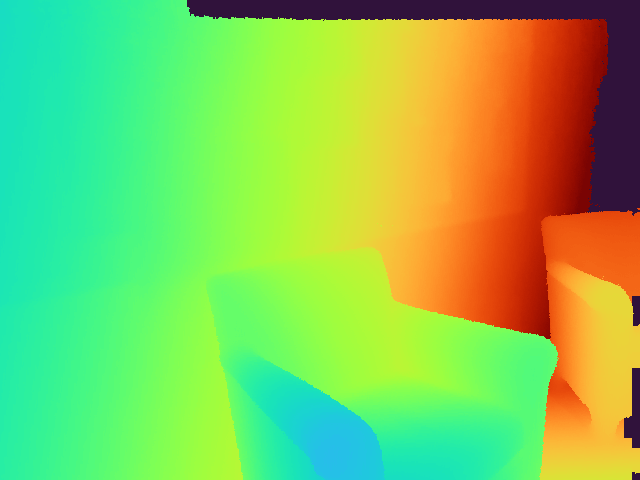} &
    \includegraphics[height = 2.2cm]{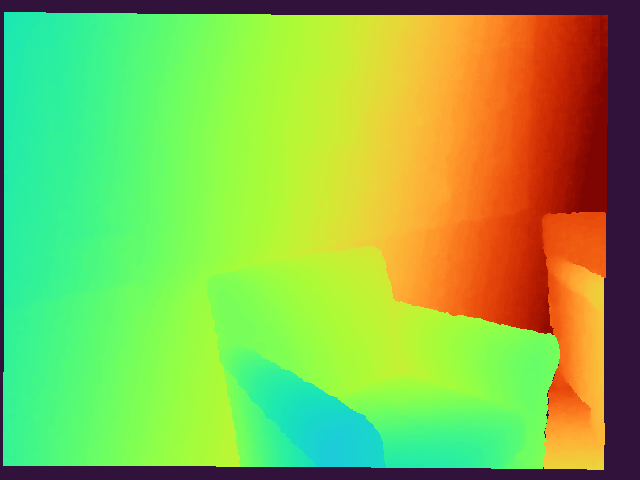}
  \end{tabular}
  \caption{Visualization of volumetric ray casting.
    From left to right: rendered depth from \emph{local}, \emph{global} hash maps, and input ground truth depth. Note the difference at boundaries.}
  \label{fig:app-raycasting-qualitative}
\end{figure}

Various accelerations can speed up ray casting. Adaptive spherical ray casting and a precomputed min-max range estimate~\cite{prisacariu2017infinitam}  will constrain the search range and boost performance. The latter can be conducted by simply projecting the active keys collected in Listing~\ref{alg:double-hash-map} without the involvement of hash maps.
In addition, we can squeeze more from our double-hash-map architecture. Conventional ray marching applies query in the \emph{global} hash map~\cite{niessner2013real,prisacariu2017infinitam,dong2019gpu}. 
Since the \emph{local} hash map is directly associated with the \emph{global} hash map with shared active indices, we can replace the \emph{global} hash map with the \emph{local} one accompanied with the $\vv^B$ buffer in the \emph{global} hash map.
With such a simple change, we can now query the more compact \emph{local} hash map, and access the \emph{global} hash map in-place without touching the geometric computations. Since out-of-frustum voxel blocks are ignored, this operation slightly sacrifices rendering completeness at image boundaries, as shown in Fig.~\ref{fig:app-raycasting-qualitative}. However, it is able to boost speed by a factor of 5, and is useful for real-time systems such as dense SLAM.

\subsubsection{Retargetable Reconstruction System}

\begin{figure*}[ht]
    \includegraphics[width=\linewidth]{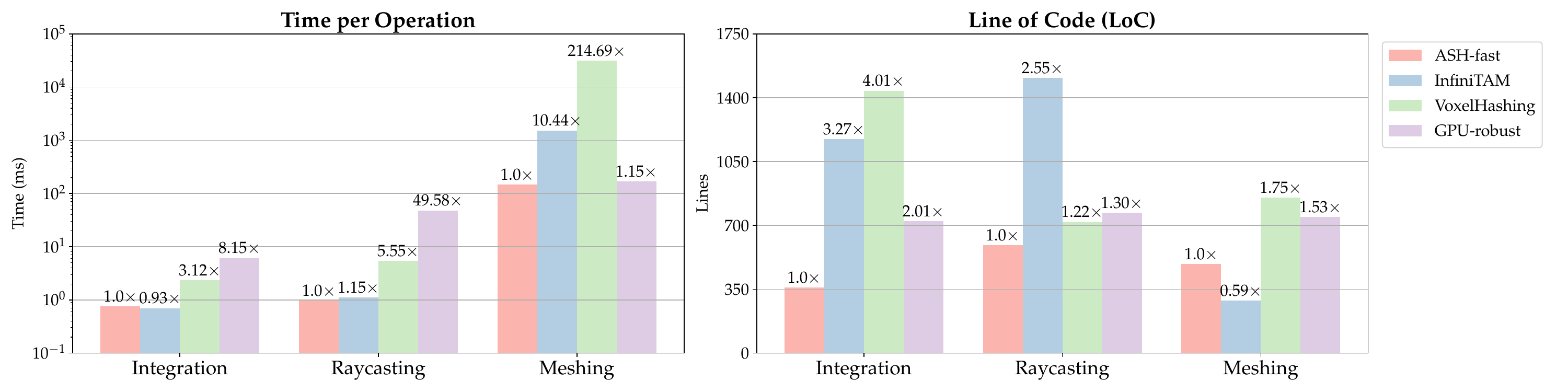}
    \caption{Performance and LoC comparison of our real-time dense SLAM pipeline in the \emph{fast} mode (ASH-fast) against state-of-the-art implementations: InfiniTAM~\cite{prisacariu2017infinitam}, VoxelHashing~\cite{niessner2013real}, GPU-robust~\cite{dong2018efficient}. Evaluated on the \emph{lounge} scene~\cite{choi2015robust}. Left: detailed comparison of separating modules. Right: corresponding LoC comparison. \emph{Lower is better}.
    Note the meshing LoC in InfiniTAM is significantly fewer since the implementation is over-simplified and requires further postprocessing. ASH-fast achieves a consistent fast speed with fewer LoC. }
    \label{fig:app-tsdf-comparison}
\end{figure*}

Thanks to the design where spatial hashing and geometry processing are detached, the spatially hashed volumetric representation can be reused for multiple purposes, from posed RGB-D surface reconstruction, dense RGB-D SLAM, to large-scale LiDAR surface reconstruction, with minimal modifications.
\begin{figure}[ht]
  \centering
    \begin{tabular}{cc}
        \includegraphics[width=0.9\linewidth]{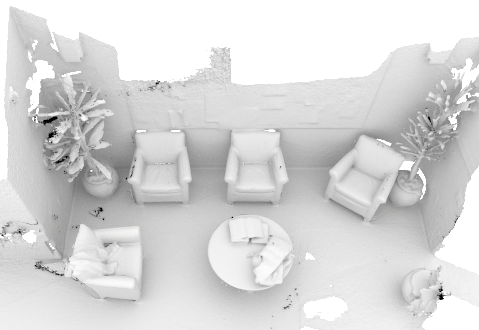} \\
        \includegraphics[width=0.9\linewidth]{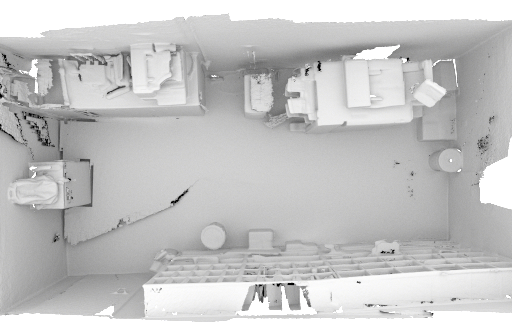}
    \end{tabular}
    \vspace{2mm}
    \caption{Visualization of triangle mesh extracted from the real-time dense SLAM system on scene \emph{lounge} and \emph{copyroom} in the \emph{fast} mode. Rendered with Mitsuba 2~\cite{nimier2019mitsuba}.}
    \label{fig:app-realtime-qualitative}
\end{figure}
For RGB-D input, a handful of existing reconstruction systems~\cite{niessner2013real,prisacariu2017infinitam,dong2019gpu} run on GPU.
We first retarget our system to \emph{fast} SLAM setup for fair comparisons. In this setup, we use a less aggressive ray-based allocation strategy~\cite{prisacariu2017infinitam} and store only weighted TSDF, consistent with the baselines. Camera poses are estimated in real-time via frame-to-model alignment~\cite{niessner2013real} between input frames and ray-casting rendering through the double-hash-map.
We compare the performance of this setup against the state-of-the-art implementations with the same parameters: voxel size is 5.8mm, voxel block resolution is $\ell = 8$, TSDF truncation distance is 4cm, min/max acceptable range of depth scanning is 0.2m and 3m.
Performance is profiled on the \emph{lounge} scene shown in Fig.~\ref{fig:app-realtime-qualitative}.
Breakdown analysis of runtime and LoC\footnote{All the code are reformatted with {\texttt clang-format} with a modified Google style.} is shown in Fig.~\ref{fig:app-tsdf-comparison}. In most comparisons, we can see a significant performance gain, with fewer LoC to write thanks to the elimination of redundant hash map look-ups in geometry kernels.

\begin{figure}[ht]
    \centering
    \begin{tabular}{@{}c@{\hspace{1mm}}c@{}}
        \includegraphics[height=3.6cm]{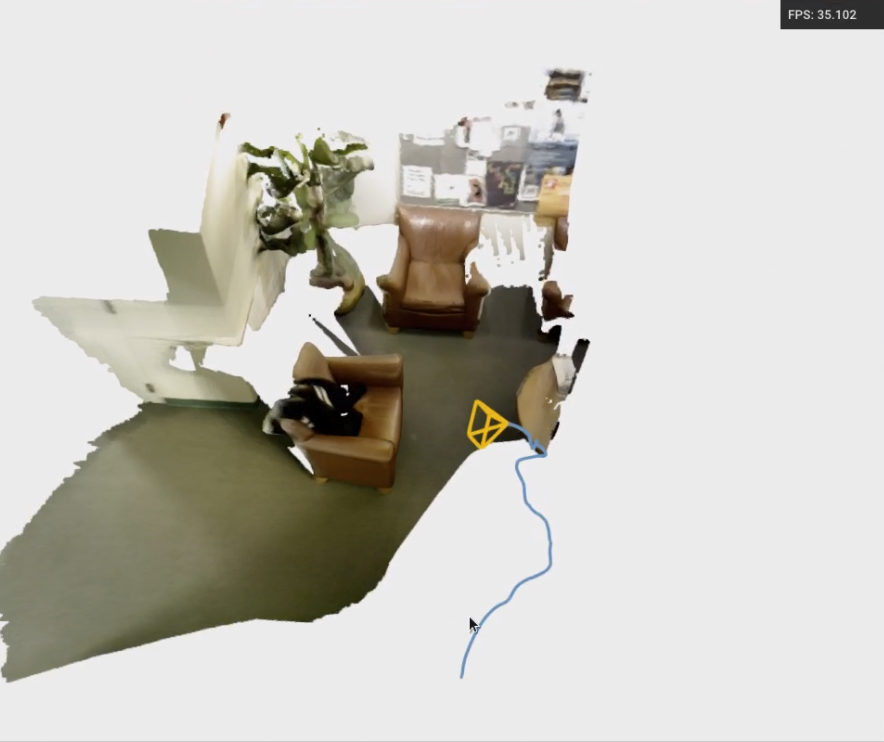}
        \includegraphics[height=3.6cm]{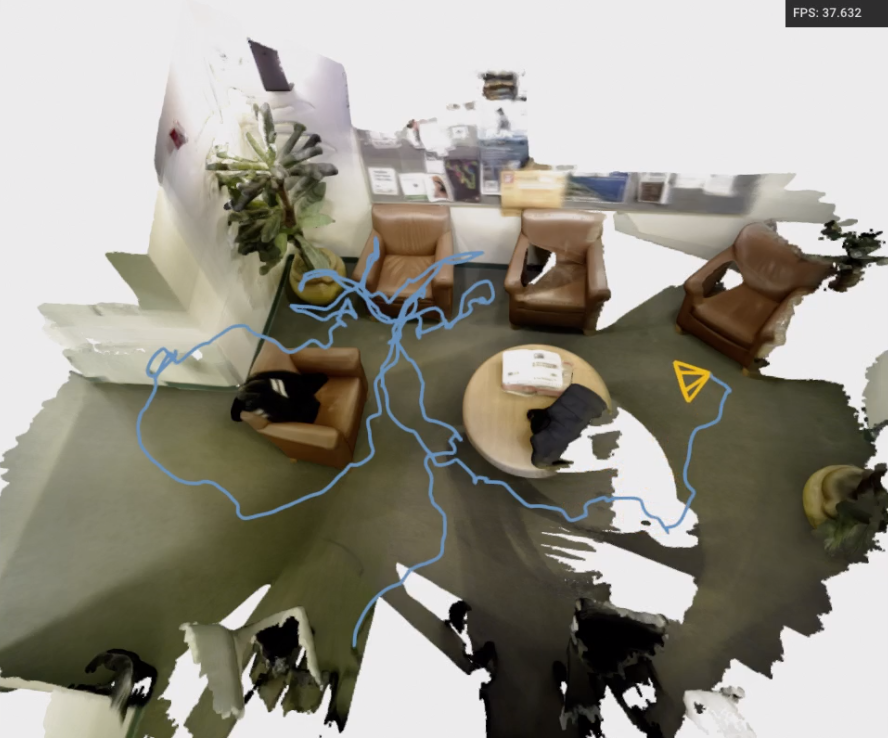}
    \end{tabular}
    \caption{Visualization of the real-time dense SLAM system in the \emph{quality} mode with colored and interactive surface reconstruction. Viewpoints can be changed by users to visualize the incremental reconstruction of the scene.}
    \label{fig:app-realtime-system}
\end{figure}

\emph{Fast} mode runs far beyond framerate and reconstructs pure geometry. In addition, we implement a \emph{quality} mode for richer volumetric information including color, and reduce noise by integrating depth into $16^3$ TSDF voxel blocks from a 1-radius-neighbor allocation~\cite{zhou2018open3d}.
The introduction of color requires double memory and triple computation cost in trilinear interpolation for ray casting and surface extraction, thus the \emph{quality} mode is around $3\times$ slower. However, it still runs in real-time, and provides a better user experience.
Fig.~\ref{fig:app-realtime-system} shows the interactive reconstruction system in the \emph{quality} mode. The system runs at 30Hz on a mid-end laptop, providing incremental volumetric reconstruction and interactive point cloud extraction and realistic rendering.
Note that to retarget from a \emph{fast} system to a user-friendly \emph{quality} application, we only need to change the block allocation function and several parameters, in total several dozen of lines, without re-writing the core.

\begin{figure}[ht]
  \centering
  \begin{tabular}{@{}c@{}}
    \includegraphics[width=0.98\linewidth, trim={0 3cm 0 0}, clip]{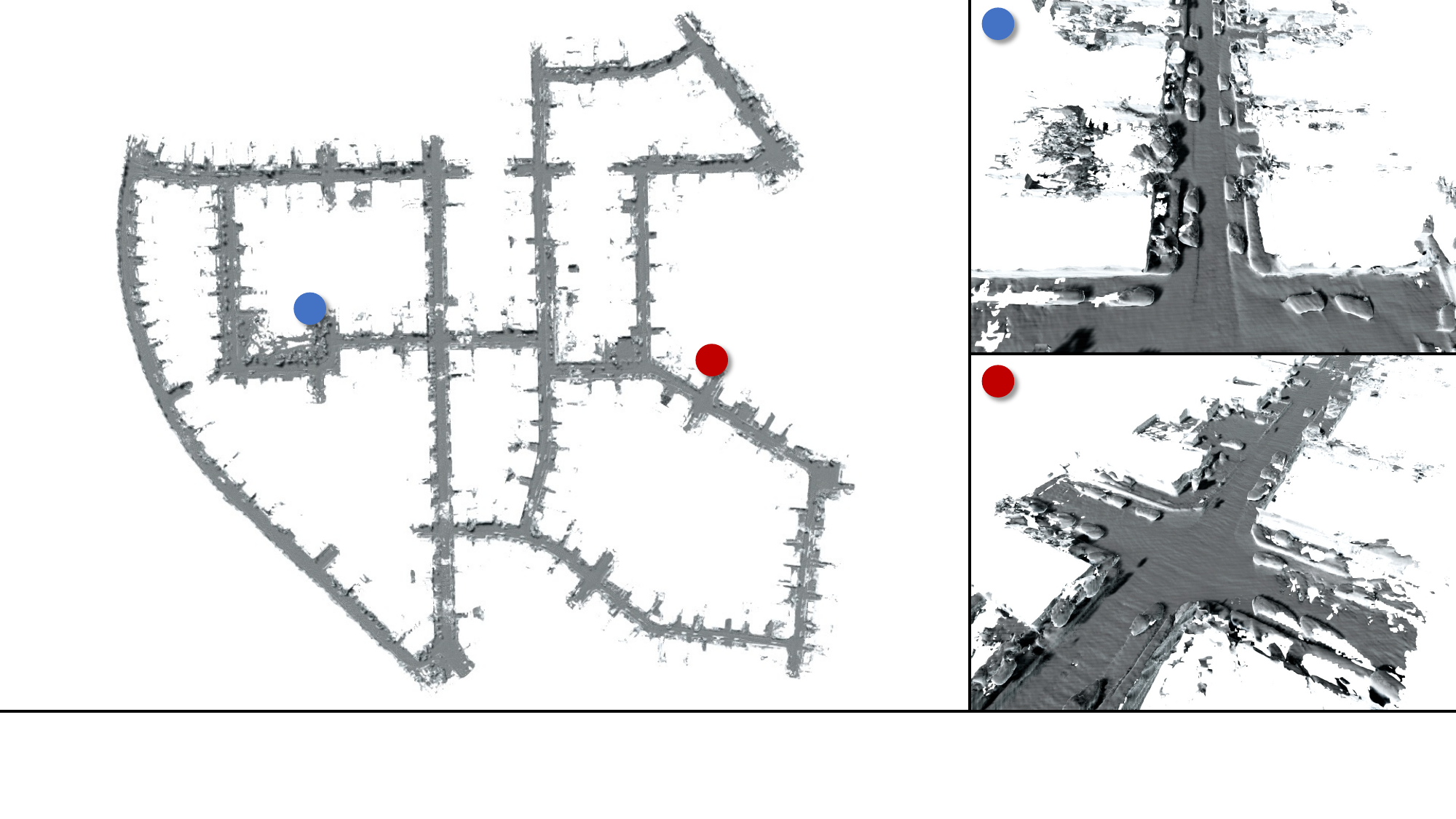} \\
    \includegraphics[width=0.98\linewidth, trim={0 3cm 0 0}, clip]{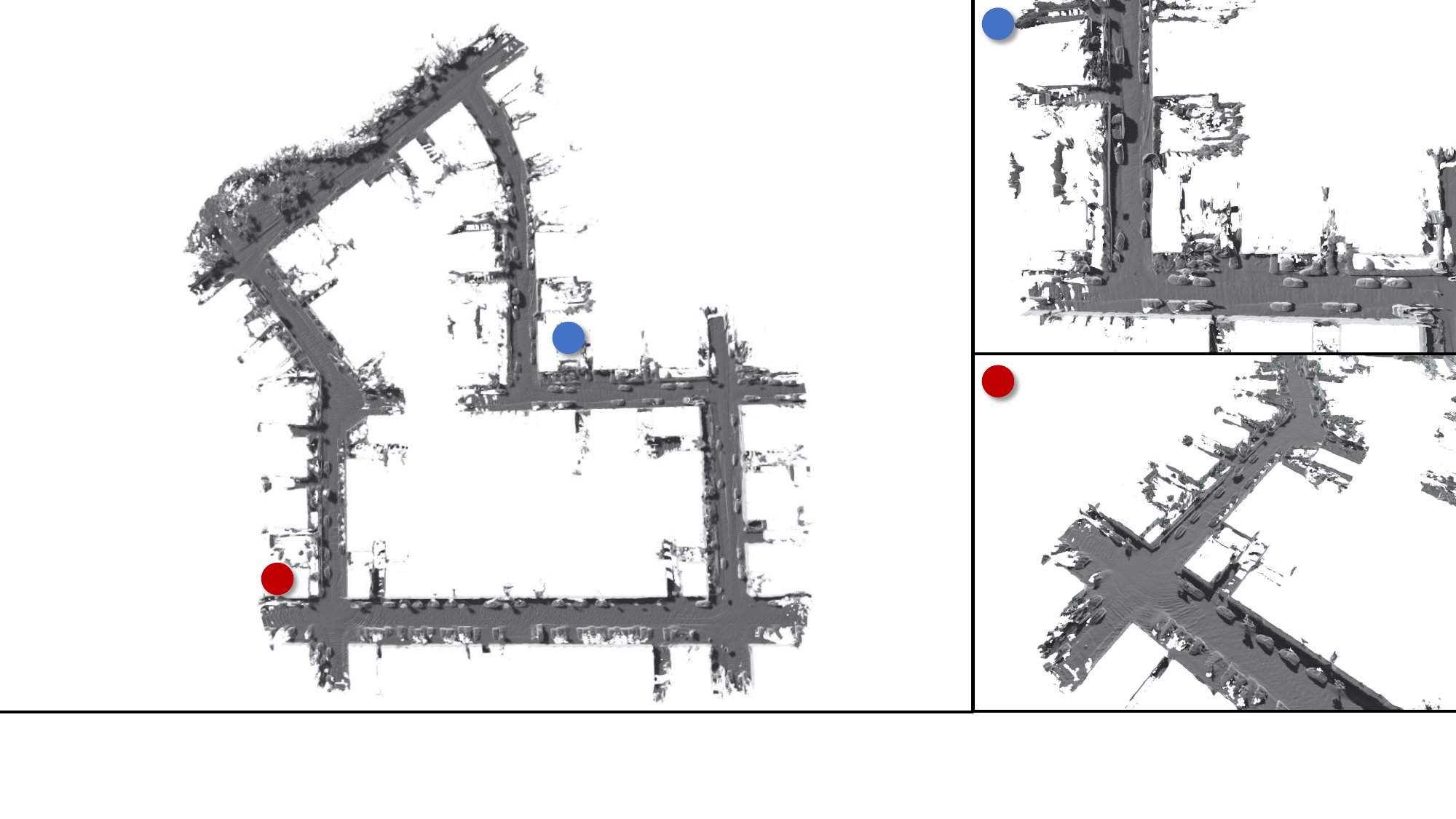}
  \end{tabular}
  \vspace{2mm}
  \caption{City-scale volumetric TSDF reconstruction of KITTI LiDAR sequence 00 and 07 with a voxel size of 0.5m.}
  \label{fig:app-kitti}
\end{figure}

The system can also be adapted to LiDAR point clouds. By simply replacing the conventional pinhole camera model with a customized spherical projection model~\cite{dong2021lidar}, we can reconstruct large-scale scenes from LiDAR data. Fig.~\ref{fig:app-kitti} shows the fast city-scale scene reconstruction from posed KITTI LiDAR point clouds~\cite{geiger2012we}. 
At such a scale, to the best of our knowledge, GPU accelerated volumetric reconstruction has not been achieved before. Thus we compare our approach against efficient CPU LiDAR volumetric reconstruction pipelines~\cite{hornung2013octomap, oleynikova2017voxblox} in the Ouster imaging LiDAR dataset~\cite{dong2021lidar}. The retargeted reconstruction system results in significant performance gain, better reconstruction quality~\cite{dong2021lidar}, at a small developing cost in LoC, as seen in Table~\ref{tab:app-lidar-loc}.
\begin{table}[ht]
    \vspace{-2mm}
    \caption{Performance and LoC (top) and reconstruction quality (bottom) comparison between ASH, OctoMap~\cite{hornung2013octomap}, and Voxblox~\cite{oleynikova2017voxblox} on the Ouster LiDAR dataset~\cite{dong2021lidar}. ASH is faster with fewer LoC, and produces a better reconstruction.}
    \centering
    \ra{1.05}
    \small
    \resizebox{0.99\linewidth}{!}{
    \begin{tabular}{lcccccc}
      \toprule
      \multirow{2}{*}{\textbf{Scenario}} & \multicolumn{2}{c}{OctoMap} & \multicolumn{2}{c}{Voxblox} & \multicolumn{2}{c}{ASH}\\
      \cmidrule(l{3mm}r{3mm}){2-3} \cmidrule(l{3mm}r{3mm}){4-5} \cmidrule(l{3mm}r{3mm}){6-7}
      & Time (ms) & LoC  & Time (ms) & LoC & Time (ms) & LoC\\
      \midrule
      Indoor & 676.20 & 25371 & 1003.26 & 13725 &  \textbf{22.29} & \textbf{2766}\\
      \midrule
      Outdoor & 1167.68 & - & 1002.22 & - & \textbf{58.55} & -\\
      \midrule
      \multicolumn{2}{l}{\textbf{Reconstruction quality}} & & \multicolumn{2}{c}{}\\
      \midrule
      F-score ($\uparrow$) &  \multicolumn{2}{c}{92.06} & \multicolumn{2}{c}{{97.98}} &  \multicolumn{2}{c}{\textbf{98.34}}\\
      \bottomrule
    \end{tabular}
    }
    \label{tab:app-lidar-loc}
\end{table}

To conclude this subsection, we presented a retargetable volumetric reconstruction system with a modular design, separating hash maps and geometric operations. With minimal changes in geometry functions, the core volumetric representation can be used for RGB-D and LiDAR scene reconstruction, and interactive dense SLAM. Our system is faster, requires fewer LoC, and supports an easy switch between speed and fidelity. 

%% file: tex/apps/slac.tex
\subsection{Non-Rigid Volumetric Deformation}
While fast online volumetric reconstruction is useful in exploration and visualization, offline reconstruction systems~\cite{choi2015robust} are sometimes preferred when a higher quality is required for design and evaluation.

State-of-the-art offline systems adopt divide-and-conquer. Long input sequences are split into smaller subsets, each yielding a submap point cloud $\mM^j$ reconstruction with less drift. In this setup, we can simply reuse the integration and surface extraction components in the previous subsection~\cite{choi2015robust, dong2019gpu}. A global pose graph is then constructed and optimized after robust registration~\cite{choy2020deep, yang2021self} of submaps. For the details, we refer the readers to state-of-the-art RGBD reconstruction systems~\cite{choi2015robust, dong2019gpu}.

\begin{figure*}[ht]
  \centering
  \begin{tabular}{@{}cc@{}}
    \includegraphics[width=0.40\linewidth]{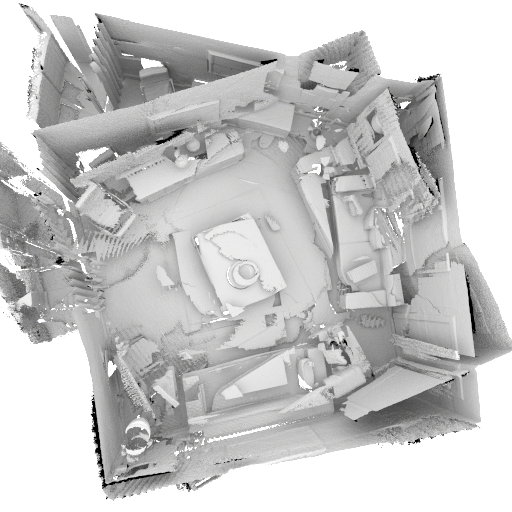} &
    \includegraphics[width=0.40\linewidth]{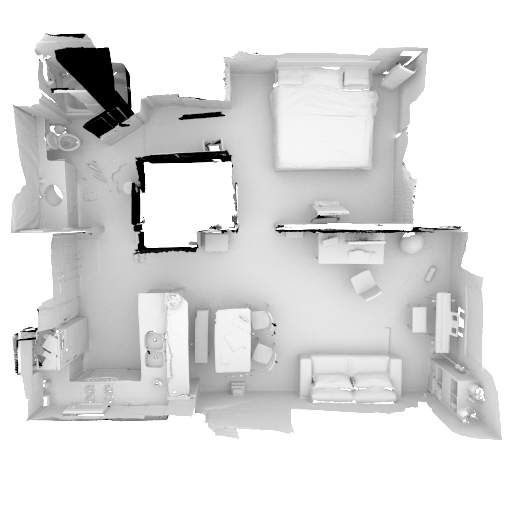} \\
    \includegraphics[width=0.40\linewidth]{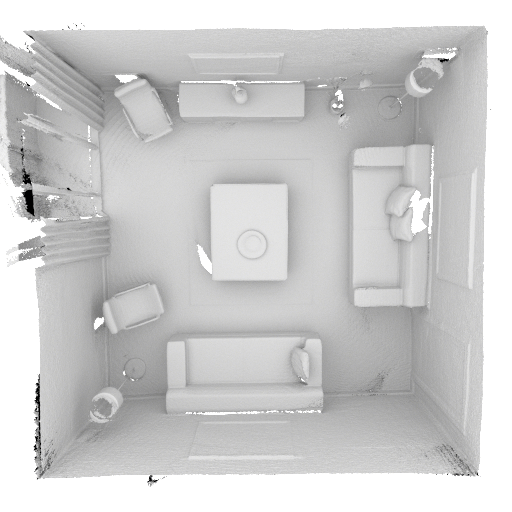} &
    \includegraphics[width=0.40\linewidth]{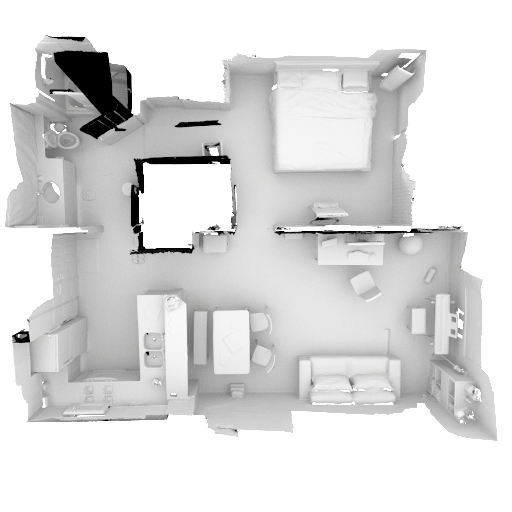}
  \end{tabular}
  \caption{Visualization of scene reconstructions before and after ASH-SLAC. First row: before ASH-SLAC. Second row: after ASH-SLAC. Left: \emph{livingroom-1} from Augmented ICL~\cite{choi2015robust}. Right: \emph{apartment} from Indoor LiDAR RGBD~\cite{park2017colored}. Artifacts are eliminated by global pose adjustment and local deformation via deformable TSDF integration. Rendered with Mitsuba 2~\cite{nimier2019mitsuba}.}
  \label{fig:app-slac-qualitative}
\end{figure*}

However, issues still persist in challenging scenes, \eg, heavy misalignment due to the strong simulated noise in the Augmented ICL dataset~\cite{choi2015robust}, and the artifacts presented in the large-scale indoor RGBD LiDAR dataset~\cite{park2017colored}, as shown in Fig.~\ref{fig:app-slac-qualitative}.
To deal with this, non-rigid volumetric deformation is presented in Simultaneous Localization and Calibration (SLAC)~\cite{zhou2013elastic,zhou2014slac}.

SLAC attempts to minimize the distance between correspondences from different submaps by optimizing a combination of rigid transformations and non-rigid deformations.
While the rigid transformations are simply the submap poses $\{\TT^j\}$, deformation is parameterized by a control grid $\cc$. In $\cc$, each grid point $\uu$ stores a local Euclidean offset $\cc_\uu \in \mathbb{R}^3$, and the accompanying function $C_\cc(\xx) = \xx + \sum_{\uu \in \NN_{\xx}} w_\uu(\xx) \cc_\uu$ deforms a point $\xx \in \mathbb{R}^3$ by applying interpolated neighbor grid offsets, where $w_\uu(\xx)$ is the interpolation ratio.
The loss function is then parameterized over $\{\TT^j\}$ and $\cc$ with as-rigid-as-possible regularizers:
\begin{align}
    \min_{\cc, \TT} &\sum_{\pp \in \mM^i, \qq \in \mM^j}\lVert(\TT^i C_\cc(\pp) - \TT^j C_\cc(\qq)\rVert^2 \nonumber \\
    + \lambda &\sum_{\uu, \vv\in \NN_\uu} \lVert \cc_\uu - \RR_\vv^{\cc_\vv} \uu \rVert^2,
\end{align}
where $\pp \in \mM^i, \qq \in \mM^j$ are corresponding 3D points between submaps obtained by nearest neighbor search. $\RR_\vv^{\CC_\vv}$ is the rigid rotation that minimizes $\lVert \RR_\vv^{\CC_\vv} (\vv - \uu) - (\cc_\vv - \cc_\uu)\rVert$ locally, where $\vv$ is a 1-ring neighbor $\NN_\uu$ of $\uu$. It controls local distortions in the as-rigid-as-possible regularizer.

\begin{figure}[ht]
    \centering
    \begin{tabular}{@{}cc@{}}
        \includegraphics[width = 0.47\linewidth]{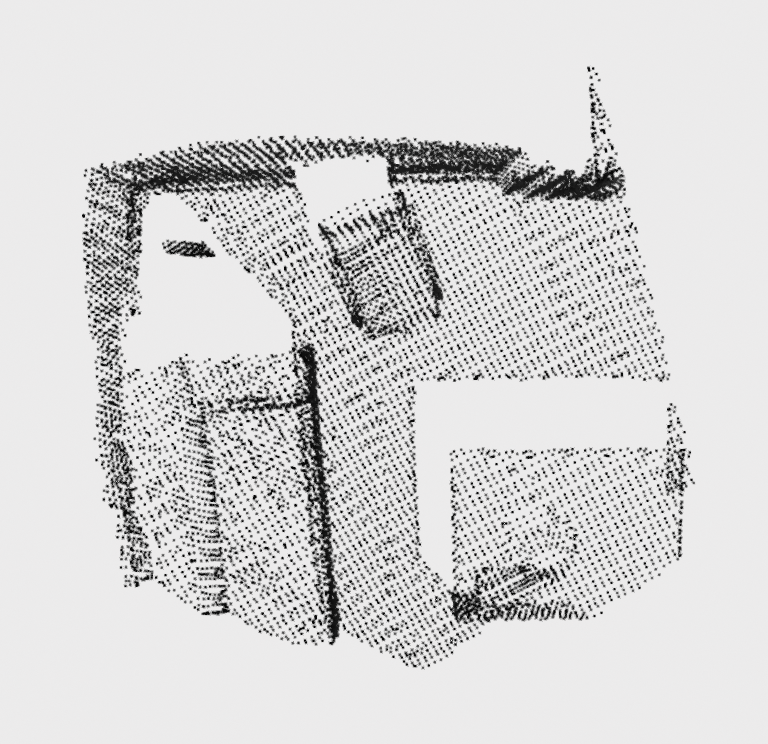}
        \includegraphics[width = 0.47\linewidth]{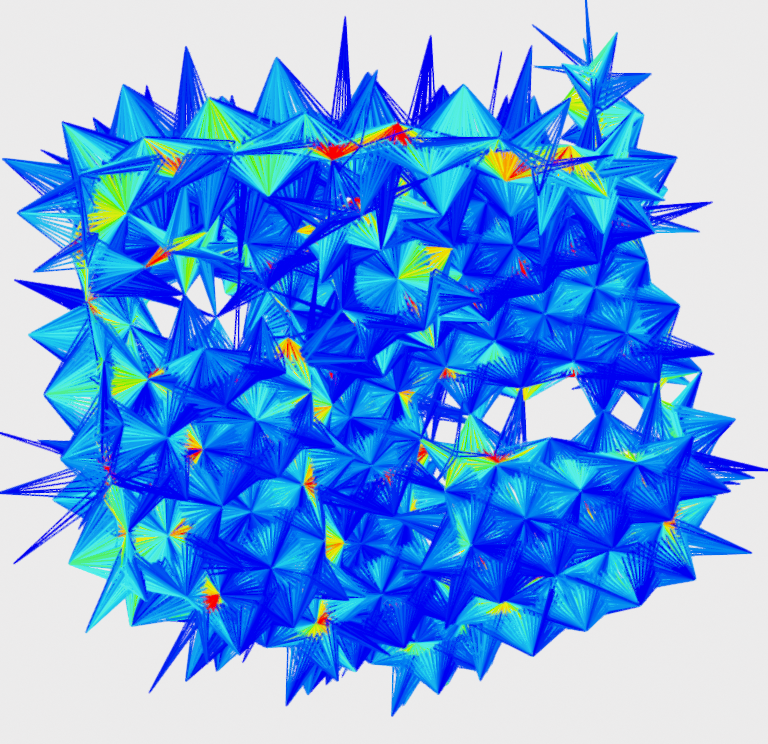}
    \end{tabular}
    \vspace{2mm}
    \caption{Visualization of a point cloud and its embedding in the volumetric deformation grids. Left: original point cloud. Right: embedding graph connecting the input points and associated deformation grid points. Each edge's color indicates the interpolation weight: blue shows a lower weight (closer to $0$), while red shows a higher weight (closer to $1$).}
    \label{fig:app-slac-illustration}
\end{figure}

This problem formulation is complicated to realize in code, and in the original implementation, the deformation grid is a simplified dense 3D array where points out-of-bound are discarded during optimization. As of today, SLAC has never been reproduced apart from the original implementation.
We observe that similar operations for TSDF grids can be applied here by ASH to generate a spatially hashed control grid.
\begin{setup}
    A volumetric deformation hash map maps grid coordinates to position offsets, and the capacity of the hash map is typically $10^3$ to $10^4$:
    \begin{equation*}
        \begin{split}
            \kK & = \{\normaltt{Tensor((3), Int32)}\}, \\
            \vV & = \{\normaltt{Tensor((3), Float32)}\}.
        \end{split}
    \end{equation*}
\end{setup}
Equipped with ASH, the non-rigid deformation can be written in several lines which results in a significant drop in LoC. We first voxelize the input point cloud with the deformation grid size following Listing~\ref{alg:voxelization}. Then, instead of the 1-radius nearest neighbors ($3^3$ entries in Listing~\ref{alg:one-radius-nn}), we look for 1-cube nearest neighbors ($2^3$), where a point is enclosed in a cube formed by grid points. The interpolation ratio can be computed jointly. We also adapt the 1-radius neighbor search to 1-ring neighbors for the regularizer.
The embedding of a submap point cloud is visualized in Fig.~\ref{fig:app-slac-illustration}, where the edges indicate the association between points to grids, and the colors show the interpolation ratio. Note that this visualization is also made easy thanks to the simple interface of ASH.

With this parameterization, we reproduce SLAC after rewriting the non-linear least squares solver and jointly optimizing the grid points and submap poses given the correspondences. In addition, the hash map can be saved and loaded from the disk for further processing, including deformed TSDF integration that reconstructs the scene from the deformed input depth images embedded in the grids.
Experiments show that with a modularized design and a spatial hash map, we can reproduce SLAC by reducing artifacts after optimization, as shown in Fig.~\ref{fig:app-slac-qualitative}.

We can see a gain in performance with fewer LoC in Table~\ref{tab:app-slac-loc} in the \emph{livingroom 1} scene with heavy simulated noise\footnote{To control the experiment, we use the initial submap pose graph from the baseline implementation.}.
Note while the hash map generalizes deformation grids from bounded to unbounded scenes, the LoC and time contributing to the core non-linear least squares optimization are slightly reduced. 
Meanwhile, the deformation and integration speed \emph{per frame} is significantly faster ($11.8\times$), which is critical for large-scale ($\ge 30$K frames) sequences. While being faster and easier to develop, our system achieves a higher reconstruction quality in terms of precision, recall, and F-score with a distance threshold $\tau = 20$mm~\cite{park2017colored}.
\begin{table}[ht]
    \vspace{-2mm}
    \caption{Performance and LoC (top) and reconstruction quality (bottom) comparison between ASH-SLAC and the original implementation~\cite{zhou2014slac} on the \emph{livingroom-1} scene~\cite{choi2015robust}. ASH-SLAC is faster with fewer LoC, and produces a better reconstruction.}
    \centering
    \ra{1.05}
    \small
    \resizebox{0.95\linewidth}{!}{
    \begin{tabular}{lcccc}
      \toprule
      \multirow{2}{*}{\textbf{Operation}} & \multicolumn{2}{c}{Original SLAC} & \multicolumn{2}{c}{ASH-SLAC}\\
      \cmidrule(l{3mm}r{3mm}){2-3} \cmidrule(l{3mm}r{3mm}){4-5}
      & Time (ms) & LoC  & Time (ms) & LoC\\
      \midrule
      Non-rigid optim. & 2041.1 & 1585 & \textbf{1982.1} & \textbf{1535} \\
      \midrule
      Deformed integration & 125.38 & 944 & \textbf{10.62} & \textbf{446} \\
      \midrule
      \multicolumn{2}{l}{\textbf{Reconstruction quality}} & & \multicolumn{2}{c}{}\\
      \midrule
      Precision ($\uparrow$) &  \multicolumn{2}{c}{29.19} & \multicolumn{2}{c}{\textbf{36.10}}\\
      Recall ($\uparrow$) &  \multicolumn{2}{c}{51.44} & \multicolumn{2}{c}{\textbf{61.34}}\\
      F-score ($\uparrow$) &  \multicolumn{2}{c}{37.24} & \multicolumn{2}{c}{\textbf{45.45}}\\
      \bottomrule
    \end{tabular}
    }
    \label{tab:app-slac-loc}
    \vspace{-5mm}
\end{table}

%% file: tex/apps/sfs.tex
\subsection{Joint Geometry and Appearance Refinement}
SLAC reduces artifacts for large-scale scenes. For small-scale objects, while volumetric reconstruction outputs smooth surfaces, fine details are often impaired due to the weight averaging of the TSDF.

\begin{figure}[ht]
    \centering
    \begin{tabular}{@{}c@{\hspace{1mm}}c@{}}
        \includegraphics[width = 0.49\linewidth]{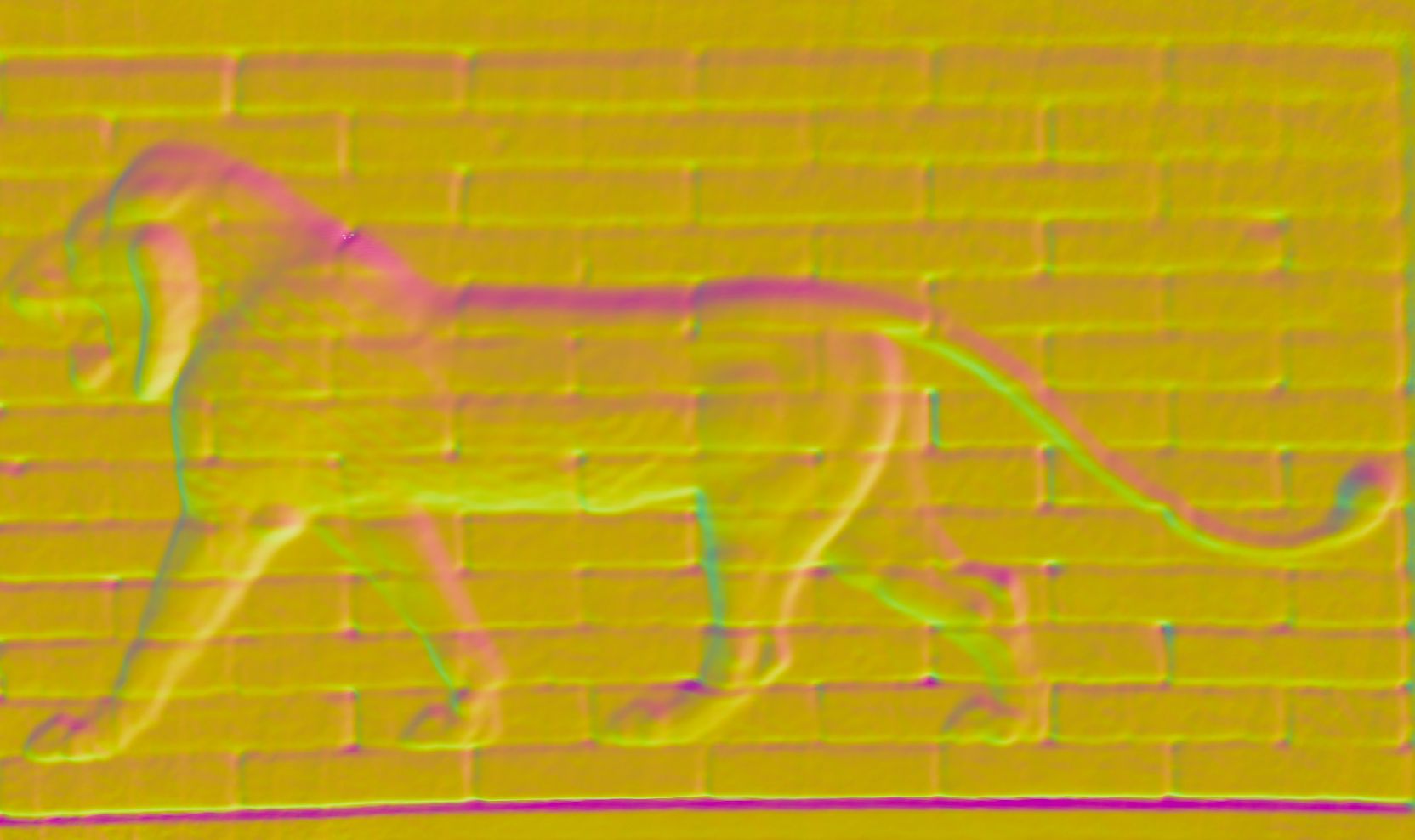} &
        \includegraphics[width = 0.49\linewidth]{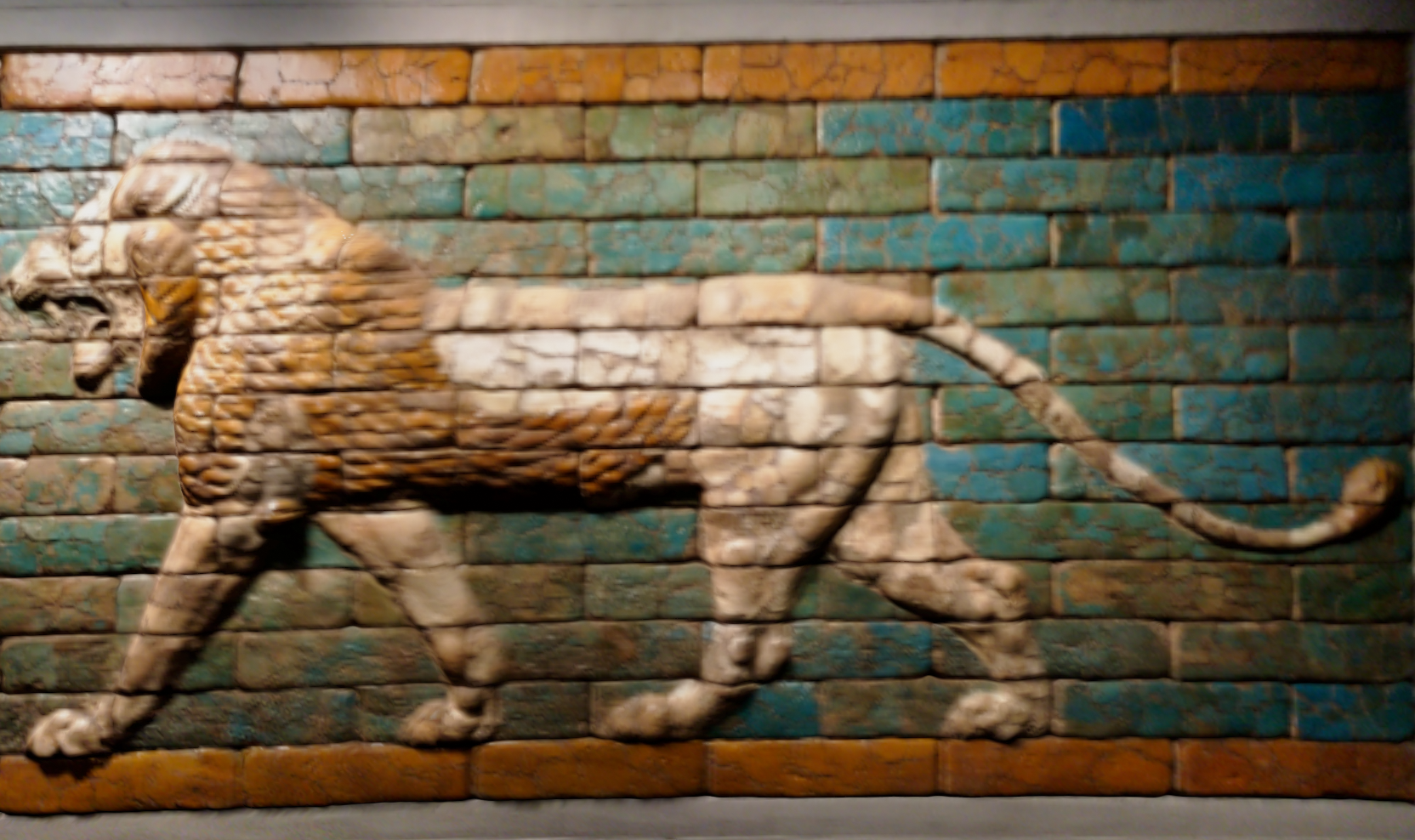} \\ 
        \includegraphics[width = 0.49\linewidth]{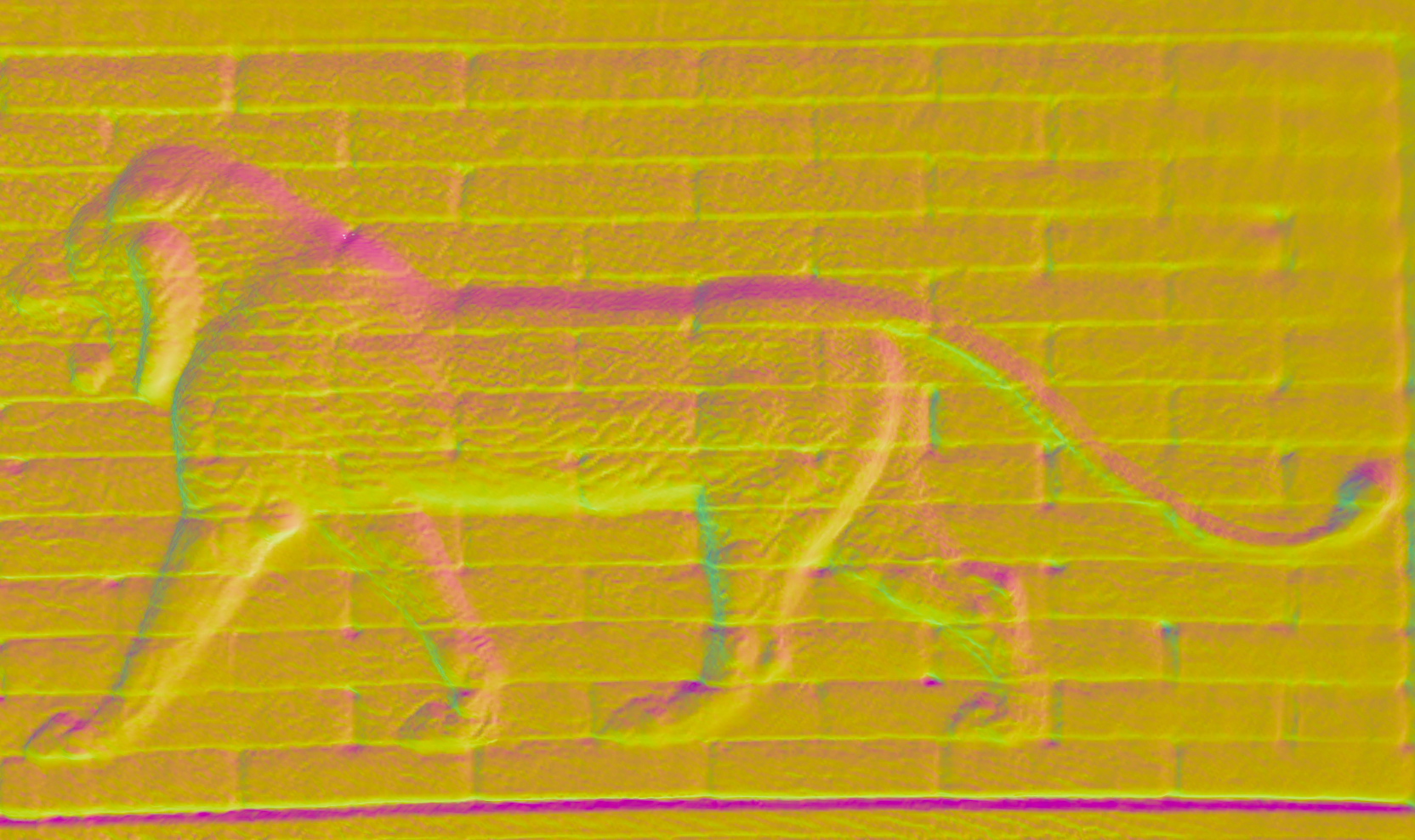} &
        \includegraphics[width = 0.49\linewidth]{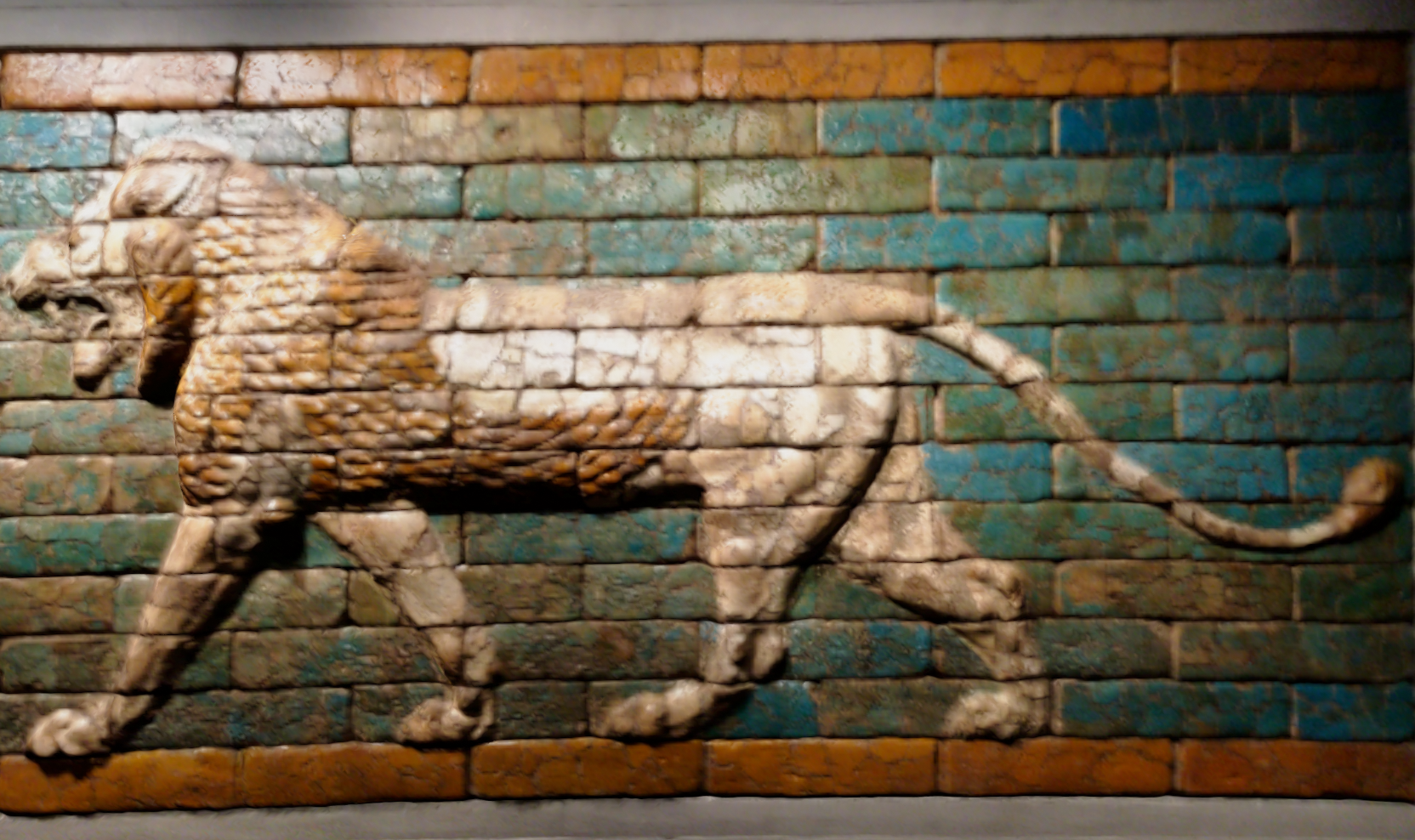}  \\
        (a) Normal map & (b) Color map
    \end{tabular}
    \vspace{2mm}
    \caption{Appearance and geometry refinement before and after ASH-Intrinsic3D on \emph{lion}~\cite{zollhofer2015shading}. First row: initial reconstruction from volumetric integration. Second row: refined reconstruction after optimization. }
    \label{fig:app-intrinsic3d}
\end{figure}

Shape-from-Shading (SfS) refines details by jointly optimizing volumetric TSDF functions given the initial geometry and appearance~\cite{zollhofer2015shading}.
It takes a reconstructed volumetric TSDF grid $\dd^0$ with a set of high-resolution key frame RGB images $I^j$ and their poses $\TT^j$ as input, and outputs jointly optimized TSDF $\dd$ and albedo $\aa$ through an image formation model
\begin{align}
    \min_{\aa, \dd} &\sum_{\xx, j}\lVert \nabla B(\xx) - \nabla I^j(\Pi({\TT^j}^{-1} \xx')) \rVert^2 \nonumber \\
    &+ \lambda_{\text{smooth}}\sum_{\xx} \lVert \Delta \dd_\xx \rVert^2 + \lambda_{\text{init}} \sum_{\xx} \lVert \dd_\xx - \dd^0_\xx \rVert^2 \nonumber \\
    &+ \lambda_{\text{chrome}} \sum_{\xx, \yy\in \NN_\xx} w(\xx, \yy)\lVert \aa_\xx - \aa_\yy \rVert^2, \label{equ:sfs}
\end{align}
where the estimated voxel-wise appearance is computed by
$B(\xx) = \aa_\xx \textrm{SH}(\nn_\xx)$ (SH stands for spherical harmonics), and associated with the closest surface point
\begin{align}
  \xx' = \xx - \dd_\xx \nn_\xx, ~~\nn_\xx = \frac{\nabla_\xx(\dd)}{\lVert\nabla_\xx(\dd)\rVert},
\end{align}
which is projected to image $I^j$ through $\Pi$ after a rigid transformation ${\TT^j}^{-1}$. Here the voxel-wise gradient is directly derived from $\dd$ with a finite difference
\begin{align}
\nabla_\xx(\dd) = \frac{\dd_{\xx+\delta} - \dd_{\xx - \delta}}{2\delta}.
\end{align}
Similar to SLAC, we use $\dd_\xx, \aa_\xx$ to access TSDF and albedo values at grid point $\xx \in \mathbb{R}^3$. $\lambda_{\text{smooth}}, \lambda_{\text{init}}, \lambda_{\text{chrome}}$ are coefficients for regularizing smoothness through the Laplacian, stability, and piece-wise albedo constancy via a weighted chromaticity regularizer $w(\xx, \yy)$, respectively~\cite{zollhofer2015shading}.

While the image formation model is straightforward, similar to SLAC, the underlying data structure used in implementing the model can be complex mainly because of the prevalent nearest neighbor search in normal computation and neighbor voxel regularizers. As a result, to enable such a system without a modern hash map, one has to rely on low-level C++ implementation and is consequently limited to the low-level Ceres solver~\cite{ceres-solver} for autodiff in optimization. Further, the spatially hashed voxels have to be bounded to reduce computation cost~\cite{maier2017intrinsic3d}.

Now equipped with ASH, we provide a simplified solution that is built upon the hash map and advanced indexing. Unlike SLAC which requires time-consuming deformable TSDF re-integration for final scene reconstruction, SfS allows reusing accelerated surface extraction from the TSDF grids without further optimization. Therefore, we implement the SfS pipeline in \emph{pure Python} as an example of fast prototyping of a differentiable rendering pipeline. Running on GPU, we lift the constraint of a user-defined bounding box and optimize the full reconstructed surface.

Without the requirement of extreme performance, we drop the hierarchical volumetric layout and use the simple voxel-based hash map:
\begin{setup}
   A voxel indexer is given by a hash set $\kK = \{\normaltt{Tensor((3), Int32)}\}$.  The typical capacity is $10^9$ to $10^{10}$.
\end{setup}
With this setup, we can reuse the code in SLAC to look up the 1-ring neighbors for normal estimation and Laplacian regularization. There is, however, another lookup required since we are minimizing the difference of appearance \emph{gradient} in Eq.~\ref{equ:sfs}: we need to find the 1-ring neighbors that also \emph{have} 1-ring neighbors. In other words, we have to find the intersection of two sets. While {\ttfamily NumPy} provides the functionality for 1D arrays through \emph{ordered} sorting, our hash map allows \emph{unordered} intersection that can be generalized to multi-dimensional inputs:
\begin{setup}
    With two input sets $\kk_1 \subset \kK, \kk_2 \subset \kK$, the intersection $\kk_1 \bigcap \kk_2$ is given by the following operations: initialize a hash set with $\kk_1$; query $\kk_2$ and obtain success mask $\btheta$; return $\kk_2(\btheta)$.
\end{setup}

After data association is found and SH parameters are estimated in a preprocessing step, all the terms in Eq.~\ref{equ:sfs} are converted to a trivial combination of indexing and arithmetic operations. We can take advantage of PyTorch's autodiff, and backpropagate the gradient through the built-in differentiable index layer. ADAM~\cite{kingma2014adam} with an initial learning rate $10^{-3}$ is used. Thus the core volumetric SfS pipeline~\cite{zollhofer2015shading} is reproduced in pure Python.

An extension can be easily implemented by introducing spatially varying lights~\cite{maier2017intrinsic3d}, wrapped up with a hash map.
\begin{setup}
    Spatially varying spherical harmonics (SVSH) (bands = 3) can be described by a hash map that maps lighting subvolume coordinates to the corresponding coefficients:
    \begin{equation*}
        \begin{split}
            \kK & = \{\normaltt{Tensor((3), Int32)}\}, \\
            \vV & = \{\normaltt{Tensor((9), Float32)}\}.
        \end{split}
    \end{equation*}
\end{setup}
The embedding of an active voxel in an SVSH map is identical to SLAC, with 1-cube neighbors for the data term and 1-ring neighbors for the regularizer. Further description is omitted here as the formulation and implementation are similar to Eq.~\ref{equ:sfs}~\cite{maier2017intrinsic3d}.

Having both SfS and SVSH optimization implemented\footnote{Pose optimization and voxel upsampling are disabled at current.}~\cite{maier2017intrinsic3d}, we show the results on the scene \emph{lion} in Fig.~\ref{fig:app-intrinsic3d}. Without voxel grid upsampling, both the geometry and appearance details are sharper. Regarding performance and code complexity, we show in Table~\ref{tab:app-sfs-loc} that our code is much shorter in pure Python, and $150\times$ faster per iteration thanks to the CUDA autodiff engine in PyTorch. Note that Ceres is a 2nd-order optimizer on CPU that empirically converges faster than the 1st-order ADAM optimizer. In practice, however, we found that in 50 iterations ADAM converges well against the preset 10 iterations for the non-linear least squares solver. Thus the total optimization performance of our implementation is still $30\times$ faster with more voxels to process (remember that we do not require an additional bounding box).

We also evaluate reconstruction quality in Table~\ref{tab:app-sfs-loc}. We render our optimized mesh given the keyframe camera extrinsic and intrinsic parameters and compute RMSE against the raw input images. For the baseline~\cite{maier2017intrinsic3d}, we follow a similar procedure and render the optimized mesh (not upsampled for fairness) given refined camera parameters. We use the same mask given by the baseline to ensure the same region of interest.
The results show that our implementation produces improved RMSE despite the simplified development.
\begin{table}[ht]
    \caption{Performance per epoch and LoC (top), and rendering quality (bottom) comparison between ASH-Intrinsic3D and the original implementation~\cite{maier2017intrinsic3d} on the \emph{lion} scene~\cite{maier2017intrinsic3d}. ASH-Intrinsic3D is faster with fewer LoC, and results in comparable rendering from the refined reconstruction.}
    \centering
    \ra{1.05}
    \small
    \begin{tabular}{lcccc}
      \toprule
      \multirow{2}{*}{\textbf{Operation}} & \multicolumn{2}{c}{Original-Intrinsic3D} & \multicolumn{2}{c}{ASH-Intrinsic3D}\\
      \cmidrule(l{3mm}r{3mm}){2-3} \cmidrule(l{3mm}r{3mm}){4-5}
      & Time (s) & LoC  & Time (s) & LoC\\
      \midrule
      SVSH optim. & 0.503 & 605 & \textbf{0.092} & \textbf{254} \\
      \midrule
      Joint optim. & 147.323 & 7399 & \textbf{0.916} & \textbf{1416} \\
      \midrule
      \multicolumn{2}{l}{\textbf{Rendering quality}} & & \multicolumn{2}{c}{}\\
      \midrule
      RMSE mean ($\downarrow$) & \multicolumn{2}{c}{0.677} & \multicolumn{2}{c}{\textbf{0.627}} \\
      RMSE std ($\downarrow$) & \multicolumn{2}{c}{\textbf{0.095}} & \multicolumn{2}{c}{0.120} \\
      \bottomrule
    \end{tabular}
    \label{tab:app-sfs-loc}
\end{table}

%% file: tex/conclusion.tex
We presented ASH, a performant and easy-to-use framework for spatial hashing. Both synthetic and real-world experiments demonstrate the power of the framework. With ASH, users can achieve the same or better performance in 3D perception tasks while writing less code.

There are various avenues for future work. At the architecture level, we seek to introduce the open address variation~\cite{alcantara2009real,junger2020warpcore} of parallel hash maps for flexibility and potential high performance static hash maps. At the low level, we plan to further optimize the GPU backend, and accelerate the CPU counterpart, potentially with cache level optimization and code generation~\cite{Enoki,hu2019taichi}.
We also plan to apply ASH to sparse convolution~\cite{choy20194d,ummenhofer2020lagrangian} and neural rendering~\cite{garbin2021fastnerf,reiser2021kilonerf}, where spatially varying parameterizations are exploited.

ASH accelerates a variety of 3D perception workloads. We hope that the presented framework will serve both research and production applications.

\section*{Acknowledgement}
We thank Patrick Stotko for adapting stdgpu to our framework, Qian-Yi Zhou for the valuable comments and detailed math derivations that helped us implement SLAC, and Kwonyoung Ryu and Jaesik Park for providing the imaging LiDAR dataset.